\documentclass[twoside,11pt]{article}
\usepackage[preprint]{jmlr2e}
\usepackage[T1]{fontenc}
\usepackage[utf8]{inputenc}
\usepackage{lmodern}
\usepackage{microtype}
\usepackage{amsmath,amssymb}
\usepackage{booktabs}
\usepackage{tabularx}
\usepackage{longtable}
\usepackage{array}
\usepackage{graphicx}
\usepackage{flafter}
\usepackage{xcolor}
\usepackage{enumitem}

\hypersetup{
  hidelinks,
  pdftitle={Action-Conditioned Transition Irreversibility for Out-of-Distribution Detection in Reinforcement Learning},
  pdfauthor={Emil Mittag, Richard Dazeley, Peter Vamplew}
}

\renewcommand{\eqref}[1]{\textup{(Equation~\ref{#1})}}

\newcolumntype{Y}{>{\raggedright\arraybackslash}X}
\newcommand{\Result}[1]{\textcolor{red!70!black}{\textbf{[result]}}}
\IfFileExists{results.tex}{\input{results.tex}}{}

\makeatletter
\def\acti@mean$#1\;#2${#1}
\def\acti@ci$#1\;#2${#2}
\newcommand{\ResultMean}[1]{\expandafter\expandafter\expandafter\acti@mean\csname actiresult@#1\endcsname}
\newcommand{\ResultCI}[1]{\expandafter\expandafter\expandafter\acti@ci\csname actiresult@#1\endcsname}
\newcommand{\ResultCell}[1]{\ifcsname actiresult@#1\endcsname
  $\ResultMean{#1}$\,{\scriptsize$\ResultCI{#1}$}\else\Result{#1}\fi}
\newcommand{\ResultCellBest}[1]{\ifcsname actiresult@#1\endcsname
  $\mathbf{\ResultMean{#1}}$\,{\scriptsize$\ResultCI{#1}$}\else\Result{#1}\fi}
\makeatother

\ShortHeadings{Action-Conditioned Transition Irreversibility}{Mittag, Dazeley and Vamplew}
\firstpageno{1}

\title{Action-Conditioned Transition Irreversibility for\\
Out-of-Distribution Detection in Reinforcement Learning}
\author{%
  \name Emil Mittag \email emil.mittag@research.deakin.edu.au \\
  \addr School of Information Technology \\
  Deakin University \\
  Geelong, Australia 3220
  \AND
  \name Richard Dazeley \email richard.dazeley@deakin.edu.au \\
  \addr School of Information Technology \\
  Deakin University \\
  Geelong, Australia 3220
  \AND
  \name Peter Vamplew \email p.vamplew@federation.edu.au \\
  \addr Institute of Innovation, Science and Sustainability \\
  Federation University Australia \\
  Mt Helen, Australia 3350
}
\date{}

\begin{document}

\maketitle

\begin{abstract}
Reinforcement-learning agents shape the observations they encounter; consequently, detecting out-of-distribution (OOD) states requires analysing action-conditioned trajectories rather than treating inputs as independent samples. We propose \textbf{Action-Conditioned Transition Irreversibility (ACTI)}, a detector that evaluates forward and backward transition likelihoods, calibrates this comparison against the in-distribution profile at each timestep, and smooths the score over a short window, using no future information. We tested ACTI across six environments and fifteen detectors within OOD-RL-Bench, using the four anomaly mechanisms that qualified in every environment, with ten policy replicates per environment. All primary metrics are matched-time: each score after anomaly onset is compared with a clean twin rollout's score at the same timestep. Our method secured the top macro matched-time AUROC, outperforming the dynamics-model baseline (PEDM), while scoring transitions roughly two orders of magnitude faster. Compared to the state-only variant, ACTI achieved a small but consistent AUROC gain and a shorter detection delay. The capacity-matched forward-only baseline, \textbf{ACTI-Forward}, secured a statistically significant advantage in AUPRC alongside a reduced detection delay. The discrepancy in AUROC did not reach statistical significance. Ablation studies suggest instead that performance improvements stem from action conditioning, calibration at each timestep, and smoothing of the transition-likelihood scores. Based on these results, we recommend using ACTI-Forward.
\end{abstract}

\begin{keywords}
out-of-distribution detection, reinforcement learning, anomaly detection, transition dynamics, time reversal
\end{keywords}

\section{Introduction}
\label{sec:introduction}

Reinforcement-learning (RL) policies are typically optimised under a restricted range of environmental conditions before being deployed into settings where sensors, actuators, dynamics, or operating regimes might change. If an agent is unable to identify these shifts, it will continue to act outside the distribution for which its policy was originally trained. While online OOD detection supplies the necessary warning mechanism, reinforcement learning presents a structural challenge distinct from detecting outliers in independent samples: specifically, the action taken at time $t$ dictates the state at $t+1$, thereby influencing all subsequent actions and observations. Consequently, the unit of analysis is not simply a state $s_t$, but an action-conditioned transition $(s_t,a_t,s_{t+1})$ embedded within a closed-loop trajectory.

Transition models provide a natural solution when deployment shifts affect dynamics rather than the marginal observation distribution \citep{danesh2021,haider2023,prashant2025}: one learns the standard mapping between $(s_t,a_t)$ and $s_{t+1}$ and then measures deviations from that pattern. However, a substantial forward-prediction residual is ambiguous. Such a residual could indicate an anomalous transition or merely reflect the inherent difficulty of prediction due to multimodal in-distribution dynamics, limited model capacity, or actions with hard-to-learn consequences. This work was motivated by investigations into fluctuation theorems regarding trajectory probabilities and time-reversal symmetry, including the time-dependent variant \citep{mittag2003}. We examine whether contrasting learned forward and inverse transition likelihoods can identify atypical transitions while reducing sensitivity to prediction challenges common to both models. This premise is assessed empirically against a capacity-matched forward-only model (ACTI-Forward).

This paper implements this detector and tests the hypothesis under controlled conditions. ACTI fits a forward model $q_{\mathrm f}(s_{t+1}\mid s_t,a_t)$ alongside a reverse model $q_{\mathrm r}(s_t\mid s_{t+1},a_t)$ using clean trajectories. It scores each step based on how much the log-likelihood contrast deviates from its per-timestep clean profile and applies causal smoothing to the result. We evaluate it in OOD-RL-Bench \citep{mittag2026} against fourteen baselines spanning distance, density, reconstruction, ensemble, sequential-change, time-series, nonlinear-dynamics, and probabilistic-dynamics families, across six environments (LunarLander, Acrobot, HalfCheetah, Ant, Humanoid, and MinAtar Breakout). The evaluation adheres to a protocol designed to eliminate the two failure modes that most commonly inflate RL-OOD results: temporal score drift (mitigated via matched-time negatives, \citealp{mittag2026}) and severity levels chosen with knowledge of detector behaviour (removed through detector-blind severity qualification). Ten independently trained policies per environment serve as the unit of replication, and the three comparisons defining the paper's primary claims: ACTI versus its state-only variant, ACTI versus the capacity-matched ACTI-Forward model, and ACTI versus PEDM \citep{haider2023} were fixed prior to observing any detector outcomes.

The key results of this study are detailed below:

\begin{enumerate}[leftmargin=*]
  \item \textbf{Benchmark performance.} ACTI achieves the highest macro matched-time AUROC among the fifteen evaluated detectors (\Result{main.matched_time_auroc.acti.macro}) and records the lowest operating-point false-positive rate (FPR at 95\% TPR, \Result{operating.matched_time_fpr_at_95tpr.acti.macro}). It maintains a statistically significant advantage over PEDM across AUROC, AUPRC, and FPR metrics, while scoring transitions in just \Result{point.score_latency.acti} compared to \Result{point.score_latency.pedm} for PEDM.
  \item \textbf{Action conditioning.} Relative to the state-only variant, action conditioning provides a consistent but modest improvement in AUROC (\Result{contrast.matched_time_auroc.acti_minus_sti}) and reduces detection delay by \Result{point.contrast_delay.acti_minus_sti_rounded} steps.
  \item \textbf{Irreversibility contrast.} The matched-capacity ACTI-Forward model was configured with the same calibration and smoothing parameters as ACTI. We observed no statistically significant divergence in matched-time AUROC or FPR at a 95\% TPR, although ACTI-Forward attained a notably better matched-time AUPRC and reduced detection delay. Consequently, these experiments failed to exhibit any performance advantages derived from contrasting forward and inverse likelihoods.
  \item \textbf{Component attribution.} At the middle severity tier $0.6$, increasing the causal smoothing window from $L=1$ to $L=20$ contributes an additional \Result{point.contrast_auroc.window_20_minus_window_1} to the AUROC and per-timestep clean-profile calibration adds \Result{point.contrast_auroc.timestep_vs_global}; across all tiers, action conditioning contributes \Result{contrast.matched_time_auroc.acti_minus_sti}. In contrast, no discernible benefit is obtained by augmenting ACTI with a joint state--action occupancy score.
  \item \textbf{Effect of the evaluation protocol.} Under within-episode (non-matched) metrics using a single-policy pilot, the performance gap between state-only and action-conditioned methods appeared to be \Result{pilot.within_episode_auroc_gap.acti_minus_sti} AUROC; however, with matched-time negatives and policy replication, this figure is \Result{point.contrast_auroc.acti_minus_sti}. Eight of the fifteen detectors in the suite incorrectly distinguish counterfactual post-onset labels on completely clean trajectories. These false detections constitute an artefact that matched-time evaluation removes.
\end{enumerate}

In our view, the mechanism-level null finding (Finding 3) and the protocol outcomes (Finding 5) hold equal importance to the benchmark result. Without capacity-matched controls and matched-time evaluation, it is probable that the detector's performance would have been attributed to the incorrect mechanism. The full component decomposition, operating analysis, and null diagnostics (tests on entirely clean trajectories, where any apparent detection is an artefact) are detailed in Sections~\ref{sec:results} and~\ref{sec:discussion}.

\section{Related Work}
\label{sec:related-work}

\subsection{OOD detection for reinforcement learning}

Danesh and Fern conceptualised OOD dynamics detection as the identification of shifts within a temporal process's dynamics and introduced RL-relevant benchmarks alongside baselines utilising recurrent prediction \citep{danesh2021}. Müller et al.\ contended that anomaly detection in RL warrants distinct treatment because sequential decision-making fundamentally alters the meaning and consequences of encountering anomalous observations \citep{mueller2022}. In parallel, Mohammed and Valdenegro-Toro benchmarked OOD methods for deep RL under conditions involving physical parameter shifts and observation corruptions \citep{mohammed2021}. Collectively, these studies underscore the necessity of evaluating detection mechanisms on trajectories rather than isolated static inputs.

Haider et al.\ introduced PEDM, a probabilistic ensemble dynamics model that identifies OOD conditions via sampled prediction error \citep{haider2023}; this represents the most robust action-conditioned baseline in our suite and serves as one of the three pre-declared comparisons. Nasvytis et al.\ proposed a formalisation of RL OOD detection, introduced scenarios involving temporal autocorrelation, and developed DEXTER, a method that extracts time-series features for application to per-dimension isolation forests \citep{nasvytis2024}. We implemented DEXTER and its sequential variant, DEXTER+C, and evaluated them alongside the other baselines. Furthermore, Prashant et al.\ defined OOD transitions based on shifts in transition probability, combining conditional variational autoencoders with conformal guarantees \citep{prashant2025}, thereby reinforcing the definition of RL-OOD through the lens of transition dynamics.

Mittag et al.~\citeyearpar{mittag2026} introduced OOD-RL-Bench, a framework developed for evaluating OOD detectors against anomalies injected into RL trajectories, and evaluated eight detectors against nine different anomalies on LunarLander using matched-time negatives. In that evaluation, a nonlinear-dynamics detector obtained a raw AUROC of $0.90$ and a matched-time AUROC of $0.54$, a difference attributed primarily to within-episode score drift. The present study employs OOD-RL-Bench to evaluate detectors using matched-time negatives, policy-level replication, and detector-blind severity qualification across six environments.

\subsection{Time reversal and irreversibility}

The quantitative link between time-reversal symmetry and entropy production stems from the fluctuation theorems of nonequilibrium statistical mechanics, which establish the probability ratio for observing entropy-consuming versus entropy-producing trajectory segments in finite systems over specific timeframes. Experimental validation has been achieved on colloidal particle trajectories \citep{wang2002}, while the time-dependent fluctuation theorem demonstrated that sets of trajectories with conjugate values of time-integrated entropy production are effectively time-reversal images of one another \citep{mittag2003}. Seara et al.~\citeyearpar{seara2021} investigated statistical irreversibility by comparing forward and time-reversed trajectory distributions. ACTI juxtaposes independently learned forward and inverse conditional likelihoods. We treat this contrast as an empirical detection statistic, without asserting that it serves as an estimator for the pathwise irreversibility measure or physical entropy production.

Barkley et al.\ formalised time-reversal symmetry within Markov decision processes and utilised feasible reverse-time transitions for data augmentation, showing performance gains in reversible systems but losses where the symmetry assumptions fail \citep{barkley2024}. Cheng et al.\ enforced time-reversal symmetry within a latent dynamics model for offline RL, employing symmetry compliance as an OOD reliability signal \citep{cheng2023}. It remains to be determined whether an explicitly action-conditioned forward/reverse likelihood contrast genuinely enhances online OOD detection across domains and anomaly mechanisms when tested against a capacity-matched forward-only control. We address this question through a direct comparison with the capacity-matched ACTI-Forward control.

\subsection{Forward dynamics and model uncertainty}
Probabilistic ensembles are widely used to represent uncertainty in learned dynamics: PETS combines probabilistic neural networks, ensembles, and trajectory sampling for model-based control \citep{chua2018}, while deep ensembles deliver scalable predictive uncertainty \citep{lakshminarayanan2017}. ACTI is not an uncertainty method; its statistic quantifies the relative support that two fitted models assign to a transition and its reversal. The ACTI-Forward ablation in Section~\ref{sec:results-contrasts} therefore serves as the definitive test for whether this statistic conveys information beyond what a single, well-calibrated forward model already provides.

\section{Problem Setting}
\label{sec:problem-formulation}

We consider a Markov decision process $\mathcal M=(\mathcal S,\mathcal A,P,R,\gamma)$ alongside a fixed deployed policy $\pi(a\mid s)$. The in-distribution (ID) training environment operates using the transition kernel $P_{\mathrm{ID}}(s'\mid s,a)$. An out-of-distribution (OOD) deployment condition arises when any component of the agent--environment loop diverges from the training setup; we specifically investigate shifts manifesting through the observation--action--transition stream, including observation corruption, action execution faults, and changes to transition dynamics (the concrete mechanisms are described in Section~\ref{sec:design-anomaly-mechanisms}).

Each episode generates a sequence $\tau=\{(s_t,a_t,r_t,s_{t+1},d_t)\}_{t=0}^{T-1}$, where $d_t$ indicates termination. In this context, $a_t$ denotes the recorded action commanded by the policy. Should an actuator fault occur, the actual action executed by the environment may differ from the command. The detector's action-conditioned inputs use $a_t$ rather than execution metadata. The detector is trained exclusively on clean trajectories and emits a causal scalar score $S_t$ using information available no later than timestep $t+1$. By applying a threshold $h$ that is calibrated on distinct clean trajectories, we generate an alarm $\hat y_t=\mathbf{1}[S_t\ge h]$. For injected anomalies, the ground-truth label transitions from zero to one at a known onset time $t^\star$.

\begin{figure}[!htbp]
  \centering
  \includegraphics[width=.85\textwidth]{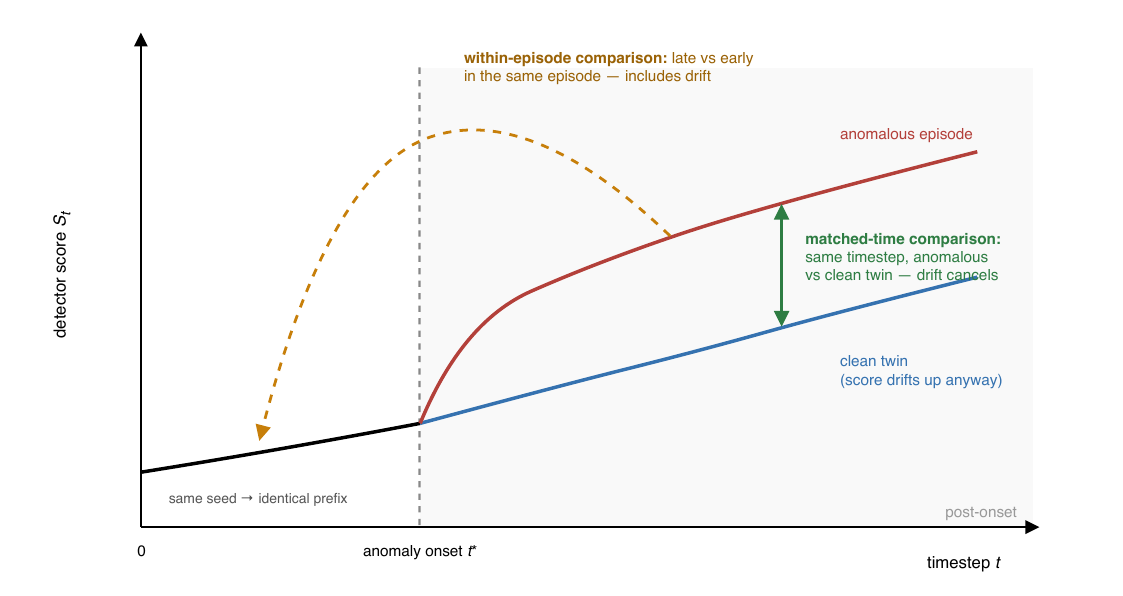}
  \caption{Matched-time evaluation. Clean detector scores naturally drift upwards during the course of an episode, even when no errors are present (blue). Comparing post-onset values against pre-onset ones within the same sequence (orange dashed arrow) essentially rewards this trend rather than detecting a fault. Instead, matched-time evaluation compares each post-onset score from the anomalous episode (red) with that of its clean counterpart—utilising the same seed and execution mode—at the corresponding timestep (green double-headed arrow). This common drift cancels out, isolating only the contribution attributable to the anomaly.}
  \label{fig:matched-time}
\end{figure}

Two characteristics of this setting drive the evaluation design. First, clean detector scores are not stationary within an episode: learned-model error shifts systematically according to the phase of the episode, meaning that comparing post-onset anomalous scores against \emph{pre-onset} scores from the same trajectory rewards drift rather than detection. Consequently, all primary metrics utilise \emph{matched-time negatives}: each post-onset anomalous score is compared against the clean score of a twin trajectory (sharing the same seed and execution mode) at the corresponding timestep \citep{mittag2026}. Specifically, regarding a horizon of 500 with an onset at $t^\star=167$, within-episode evaluation assesses whether the anomalous episode's scores during the post-onset phase ($t > 167$) exceed its own baseline from the pre-onset period ($t \le 167$). Because learned-model errors typically rise later in a sequence—particularly near terminal phases or contacts—this method may report a detection purely due to this background trend, even if the anomaly has no effect. Conversely, matched-time evaluation compares the score at time $t=300$ in the anomalous episode against the score at the identical time point ($t=300$) in its clean counterpart. This approach eliminates variance caused by the natural progression of the episode, ensuring that only the anomaly's contribution remains. Figure~\ref{fig:matched-time} illustrates both comparisons. We quantify the artefact eliminated by this approach in Section~\ref{sec:results-nulls}. Second, anomaly severity parameters exert environment-dependent, and occasionally non-monotone, effects; selecting severity levels after examining detector outcomes would introduce a selection bias. Therefore, the protocol detailed in Section~\ref{sec:design-anomalies} establishes severity levels based solely on environmental factors, remaining blind to all detector performance.

\section{ACTI and its Variants}
\label{sec:acti}

\subsection{Forward and reverse transition models}
ACTI comprises two learned one-step dynamics models fitted to clean trajectories. The \emph{forward} model determines how the state should evolve given the current state and action taken, while the \emph{reverse} model addresses the inverse problem: identifying the backward step required to recover the previous state from the resulting state and action label. Each model is assessed by how well its prediction aligns with the observed transition; comparing these fits generates the detection signal for the next subsection.

Both models estimate state \emph{increments} rather than raw successor states. Let $\Delta s_t=s_{t+1}-s_t$ be the increment computed on a real-valued state representation—the native state vector, or the output of a fixed projection for image-based observations (Section~\ref{sec:design-environments}). The forward model predicts the increment; the reverse model predicts its negation—effectively stepping from $s_{t+1}$ back to $s_t$:
$$ \widehat{\Delta s}^{\mathrm f}_t=f_\theta(s_t,a_t), \qquad \widehat{-\Delta s}^{\mathrm r}_t=g_\phi(s_{t+1},\bar a_t). $$
The residuals, $e^{\mathrm f}_t=\Delta s_t-f_\theta(s_t,a_t)$ and $e^{\mathrm r}_t=-\Delta s_t-g_\phi(s_{t+1},\bar a_t)$, quantify the deviation of the observed transition from each model's prediction. We treat these residuals as zero-mean Gaussians, $e^{\mathrm f}_t\sim\mathcal N(0,\Sigma_{\mathrm f})$ and $e^{\mathrm r}_t\sim\mathcal N(0,\Sigma_{\mathrm r})$, with covariances estimated via shrinkage on in-distribution residuals. This effectively transforms each model into a transition likelihood—the $q_{\mathrm f}$ and $q_{\mathrm r}$ discussed below. Both models consist of two-layer MLPs with matching widths, trained under identical resource budgets. Discrete actions are one-hot encoded, while continuous actions are normalised based on the statistics of the training data.

Defining the reverse action input $\bar a_t$ necessitates establishing a specific convention. As physically reversed control is rarely applicable within RL systems, this study adopts the \emph{conditional descriptor} convention ($\bar a_t=a_t$) throughout: the reverse model assesses the reversibility of the transition strictly \emph{conditional on the provided action label}. Consequently, the output serves as a proxy for statistical transition irreversibility rather than a measure of thermodynamic entropy production (refer to Section~\ref{sec:discussion-thermo}).

\subsection{Contrast, calibration, and smoothing}
The one-step action-conditioned contrast is
\begin{equation}
  i_t=
  \log q_{\mathrm f}(s_{t+1}\mid s_t,a_t)
  -
  \log q_{\mathrm r}(s_t\mid s_{t+1},\bar a_t),
\end{equation}
the difference of residual log-densities including covariance normalisation terms. Because anomalies need not increase the contrast, and need only move it away from its normal value, the score measures deviation from the clean per-timestep profile,
\begin{equation}
  z_t=
  \frac{|i_t-\mu_{\mathrm{ID}}(t)|}
  {\max(\sigma_{\mathrm{ID}}(t),\epsilon)},
  \label{eq:calibration}
\end{equation}
where $\mu_{\mathrm{ID}}(t)$ and $\sigma_{\mathrm{ID}}(t)$ are estimated per timestep from a clean profile split, guarding against normal within-episode score drift, and $\epsilon=10^{-6}$ ensures numerical stability. The online score is a causal average of the last $L$ calibrated contrasts,
\begin{equation}
  S^{\mathrm{ACTI}}_t=
  \frac{1}{\min(L,t+1)}
  \sum_{j=0}^{\min(L-1,t)}z_{t-j},
  \label{eq:smoothing}
\end{equation}
with $L=20$ fixed in advance; because smoothing trades latency for separability, Section~\ref{sec:results-components} reports the full accuracy--latency frontier over $L\in\{1,5,10,20,50\}$.

\subsection{Variants and controls}
\label{sec:acti-variants}

We evaluate three additional variants alongside forward-only (ACTI-Forward) and reverse-only controls to isolate each design choice; all configurations utilise identical calibration \eqref{eq:calibration} and smoothing \eqref{eq:smoothing} logic:

\begin{itemize}[leftmargin=*]
  \item \textbf{STI} (state-only): This mirrors the primary construction but excludes actions ($a_t$ and $\bar a_t$) from both models. Comparing ACTI against STI tests the efficacy of action conditioning.
  \item \textbf{ACTI-Forward} (forward-only): The raw statistic here relies solely on $-\log q_{\mathrm f}(s_{t+1}\mid s_t,a_t)$—without a reverse model or contrast—and employs matched architecture and training. This condition tests whether contrasting forward and inverse likelihoods enhances detection performance relative to using forward likelihood exclusively, while applying identical calibration and smoothing protocols.
  \item \textbf{reverse-only}: Symmetrically, this variant relies purely on the reverse likelihood.
  \item \textbf{Occupancy-only}: Under clean operation, the policy remains within a restricted region of the joint state--action space---its \emph{occupancy}. This variant concatenates the normalised state with the encoded action and fits the mean and shrinkage-estimated covariance of these vectors using clean trajectories. It then scores each pair $(s_t,a_t)$ based on its Mahalanobis distance relative to this estimate. It therefore targets a detection gap that transition models cannot observe: an agent displaced into unfamiliar state--action territory where individual transitions remain perfectly predictable.
  \item \textbf{ACTI+O}: This combines calibrated ACTI and occupancy scores with equal weighting to verify if the combination provides additional information beyond what occupancy metrics supply alone.
\end{itemize}

\noindent We collectively refer to STI, ACTI, and ACTI+O as the transition-irreversibility family.

While ACTI requires fitting two models compared to the single model used by ACTI-Forward, its scoring phase involves two model evaluations plus calibration with constant memory overhead. Fit time, scoring latency, throughput, and model size are reported alongside detection metrics in Section~\ref{sec:results-compute}.

\section{Experimental Design}
\label{sec:experimental-design}

\subsection{Environments and policies}
\label{sec:design-environments}

Table~\ref{tab:environments} outlines the six environments utilised in this study. LunarLander and Acrobot provide a compact setting for discrete-action control, whereas HalfCheetah, Ant, and Humanoid introduce continuous actions, contact dynamics, and escalating state dimensions. MinAtar Breakout (MinAtar \citealp{young2019}) is utilised to confirm that the detection signal remains viable when observations consist of pixels rather than low-dimensional physical state vectors, such as positions and velocities. Four stacked $10{\times}10{\times}4$ binary frames are mapped into 128 dimensions via random Gaussian projection, yielding a real-valued representation where state increments act as continuous variables and covariances can be estimated from the available clean episodes—properties that raw ${\sim}1{,}600$ binary pixels lack. This projection is data-independent, static prior to data collection, and shared across all detectors; consequently, it functions identically for both model fitting and anomalous evaluation, ensuring no detector gains an advantage through representation choices. Episodes proceed for a horizon of 500 steps in LunarLander and Acrobot, or 1000 steps in the remaining environments, with anomaly onset defined at one third of the total horizon duration.

\begin{table}[!htbp]
  \centering
  \small
  \begin{tabularx}{\textwidth}{@{}lYlYY@{}}
    \toprule
    Environment & Observation & Action & Policy & Role \\
    \midrule
    LunarLander-v3 & 8-D state & 4 discrete & DQN \citep{mnih2015} & Discrete-action control \\
    Acrobot-v1 & 6-D state & 3 discrete & DQN & Discrete-action transfer \\
    HalfCheetah-v5 & 17-D proprioception & 6-D continuous & TD3 \citep{fujimoto2018} & Regular locomotion \\
    Ant-v5 & 27-D proprioception & 8-D continuous & TD3 & Contact-rich control \\
    Humanoid-v5 & 45-D proprioception & 17-D continuous & TD3 & Model-capacity stress \\
    Breakout (MinAtar) & $10{\times}10{\times}16$ binary $\to$ 128-D projection & 3 discrete & DQN (convolutional) & Visual representation stress \\
    \bottomrule
  \end{tabularx}
\normalsize
  \caption{Environments and policies. Ten independently seeded policies per
  environment are trained and frozen before any detector fitting; the policy
  checkpoint, not the episode, is the unit of replication.}
  \label{tab:environments}
\end{table}

All environments are executed via Gymnasium~\citep{towers2024}. For each environment, we trained ten policies (using training seeds 106--115) according to a fixed recipe before freezing them; no seed selection or replacement occurred based on performance metrics. For each policy, we gathered distinct sets of clean trajectories through rollouts: 30 episodes for detector fitting, 20 for per-timestep profile estimation, another 20 for threshold calibration, and a final 20 for matched-time testing. To ensure the sets remained disjoint, we utilised separate random seeds for the rollout generation in each split. Evaluation involved applying exploration noise—5\% random actions for discrete policies and Gaussian noise with $\sigma=0.1$ for continuous ones—and introducing small calibrated process noise across the MuJoCo tasks that was applied consistently to both clean and anomalous rollouts.

\subsection{Anomaly mechanisms}
\label{sec:design-anomaly-mechanisms}

The candidate anomalies are adapted from the nine mechanisms of OOD-RL-Bench \citep{mittag2026}. Each includes a raw severity parameter $\sigma\in[0,1]$ that controls the perturbation magnitude in mechanism-specific units; the method for mapping these raw severities to the evaluated tiers is detailed in Section~\ref{sec:design-anomalies}. These mechanisms fall into two distinct execution modes. Five operate \emph{post hoc}: a clean rollout is stored, and an injector subsequently modifies the recorded observation stream, which remains feasible provided the anomaly can be represented as a corruption of the data collected. Conversely, four operate \emph{online}, being applied directly during live environment interactions because they alter the closed-loop dynamics themselves; applying these \emph{post hoc} would result in the recorded states becoming inconsistent with the actions actually executed.

\paragraph{Gaussian state noise (post hoc).}
Independent Gaussian noise is added to every state coordinate at each post-onset step. The per-coordinate standard deviation scales by $\sigma$ relative to the corresponding value in the clean trajectory; coordinates exhibiting minimal variation therefore undergo proportionately smaller perturbations. This scales the perturbation relative to each coordinate's clean variability, modelling sensor noise. It does not ensure that signs or magnitudes are preserved because Gaussian noise is unbounded.

\paragraph{State-feature dropout (post hoc).}
Each post-onset state coordinate is independently set to zero with probability $0.8\sigma$, while other coordinates pass through unaltered. This mechanism models intermittent sensor failure or telemetry loss. Since clean data rarely contains exact zeros for these coordinates, the resulting joint states fall significantly outside the fitted in-distribution support.

\paragraph{Correlation break (post hoc).}
The injector selects $\max(1,\lfloor\sigma d_s\rfloor)$ of the $d_s$ state dimensions and permutes their values across time within the post-onset window. While each affected dimension retains its marginal distribution, the temporal ordering and inter-coordinate coupling are destroyed. This simulates sensor cross-talk or a fault that incorrectly maps readings to signals.

\paragraph{Regime switch (post hoc).}
At onset, the post-onset state distribution shifts by $2.0\sigma$ pre-onset standard deviations per coordinate, while the per-coordinate variance increases by a factor of \((1+2.0\sigma)^2\). The state sequence is not re-propagated under the policy; rather, the marginal observation distribution simply transitions abruptly to a new operating regime.

\paragraph{Observation delay (post hoc).} Post-onset observations are replaced by the state from $k$ steps prior ($k=\lceil\sigma\cdot k_{\max}\rceil$, with $k_{\max}=10$), leaving the recorded actions unchanged. This simulates communication or processing latency within the observation stream. An observation from a few steps earlier can still look plausible on its own.

\paragraph{Stuck actuator (online).} Following onset, for discrete actions there is a probability $\sigma$ at each step that the executed action is replaced by the most recent one; at $\sigma=1$, the actuator maintains the onset action for the remainder of the episode. For continuous actions, a severity-dependent subset of action coordinates remains fixed at its pre-onset values, while the remaining ones follow the current command. The trajectory's action channel records only the \emph{commanded} action, with the actual executed action preserved as metadata—mirroring a deployment scenario where the fault remains invisible to standard logs. Consequently, the state trajectory diverges from its clean counterpart due to the closed-loop consequences of repeatedly executing an unintended action.

\paragraph{Dynamics scaling (online).} After onset, all state increments ($s_{t+1}-s_t$) are scaled by a constant factor of $1+1.5\sigma$ and fed back into the simulator prior to the next policy action. This models an abrupt shift in physical properties such as effective mass, friction, or actuator gain. Because subsequent actions rely on these modified states, the trajectory is driven into regions of state space that were not encountered during training.

\paragraph{Action-conditioned dynamics (online).} Increments following onset are attenuated by a factor of $1-0.8\sigma$, applied strictly only to transitions meeting a designated action condition. For discrete spaces, this implies equality with a fixed target action; for continuous spaces, it involves crossing a threshold on a specific dimension. The target is determined in advance according to the clean-support rule in Section~\ref{sec:design-anomalies}. While marginal detectors encounter only intermittent anomalies, a detector that utilises action information yields a coherent signal exactly when the condition is met, serving as a direct test of action conditioning.

\paragraph{Gradual drift (online).} The dynamics scaling factor increases linearly from $1.0$ at onset to $1+1.5\sigma$ across 200 steps in LunarLander and Acrobot and 400 steps in Breakout before remaining constant, effectively simulating progressive mechanical degradation. Because the immediate post-onset phase closely mimics clean dynamics, detectors dependent on instantaneous outliers only respond once the ramp effect accumulates sufficiently.

\subsection{Detector-blind severity qualification}
\label{sec:design-anomalies}

Severity parameters lack universal meaning across different environments, and their impact does not necessarily scale monotonically. Consequently, each environment-anomaly combination underwent \emph{qualification} prior to any detector testing. We estimated a scalar \emph{realised-effect} metric across a severity grid spanning twenty reset seeds using a pilot policy. Computed directly from trajectories for each anomaly channel, this detector-free measure quantified the extent of data alteration (e.g.\ the post-onset state RMSE relative to the matched clean counterpart, or the discrepancy between commanded and executed actions). We employed an isotonic regression—a monotone step-function fit of the realised effect against severity—to identify the distinct levels actually attainable (the plateaus). This enabled us to define five ordinal effect tiers for each pair at approximately 20/40/60/80, and 100\% of that specific pair's maximum achievable effect. These tier codes represent ordinal positions rather than calibrated percentages, and severity selection remained entirely independent of detector performance. Combinations that failed qualification were excluded along with their respective reasons. Anomalies were excluded when the realised effect significantly decreased as severity increased, or if fewer than five distinct positive plateaus emerged despite exhaustive grid refinement. Additionally, exclusions occurred if the anomaly activation condition fired on less than 20\% or more than 80\% of clean transitions for any of the five policies, or if high-severity candidates proved numerically unstable. Following the initial qualification outcomes but prior to any detector evaluation, one procedural adjustment was implemented: the minimum number of mechanisms required per environment was reduced from six to four after the original threshold proved unachievable; however, the individual per-cell qualification decisions remained unchanged.

Table~\ref{tab:qualification-map} presents the outcomes: of the 54 environment-anomaly combinations, \Result{support.qualified_cell_count} passed qualification while \Result{support.non_admitted_cell_count} did not. The common support $A^*$, defined as the set of mechanisms admitted in every environment, contains \Result{support.common_support_count} anomalies (\Result{support.common_anomaly_set}), comprising three post-hoc types and one live mechanism. All confirmatory comparisons and macro summaries utilise only $A^*$, with detailed results for the fifteen environment-specific qualified cells reported separately in Section~\ref{sec:results-environments}. Observation delay and action-conditioned dynamics, which \citet{mittag2026} reported as difficult to detect on LunarLander, also proved most difficult to control during qualification, failing in four out of six environments each.

\begin{table}[!htbp]
  \centering
  \small
  \begin{tabular}{@{}lcccccc@{}}
    \toprule
    Anomaly & LL & AC & HC & HU & BR & AN \\
    \midrule
    Gaussian state noise & P & P & P & P & P & P \\
    State-feature dropout & P & P & P & P & P & P \\
    Regime switch & P & P & P & P & P & P \\
    Stuck actuator & P & P & P & P & P & P \\
    Correlation break & S & ---$^{p}$ & S & S & S & S \\
    Observation delay & S & ---$^{r}$ & ---$^{r}$ & ---$^{r}$ & S & ---$^{r}$ \\
    Dynamics scaling & S & S & ---$^{n}$ & ---$^{n}$ & S & ---$^{n}$ \\
    Gradual drift & S & S & ---$^{n}$ & ---$^{n}$ & S & ---$^{n}$ \\
    Action-conditioned dynamics & ---$^{c}$ & S & ---$^{p}$ & ---$^{p}$ & S & ---$^{c}$ \\
    \bottomrule
  \end{tabular}
\normalsize
  \caption{Qualification map: which anomaly mechanisms were admitted per environment. LL = LunarLander, AC = Acrobot, HC = HalfCheetah, HU = Humanoid, BR = Breakout, AN = Ant; P = admitted, in the common support $A^*$ used for all primary comparisons; S = admitted in this environment only (reported as expanded support, never entering primary rankings); --- = not admitted, with reason: $^{r}$ supported effect reversal, $^{p}$ too few positive effect plateaus, $^{c}$ clean action-support failure, $^{n}$ numerical instability of high-severity candidates.}
  \label{tab:qualification-map}
\end{table}

\subsection{Detector baselines}
\label{sec:design-baselines}

The collection is composed of fifteen detectors spanning the distance, density, reconstruction, ensemble, sequential-change, time-series, nonlinear-dynamics, and probabilistic-dynamics families. Four of these (STI, ACTI, occupancy-only, and ACTI+O) are introduced in Section~\ref{sec:acti}, while the remaining eleven baselines are detailed below.

\paragraph{Mahalanobis distance \citep{lee2018}.}
This method fits the mean and ridge-regularised covariance of clean states, scoring each observation by its covariance-normalised distance from that mean. As a straightforward marginal outlier detector, it triggers when observations are displaced from the fitted in-distribution support but remains insensitive to anomalies that preserve the marginal state distribution.

\paragraph{KDE density.}
This approach estimates a non-parametric kernel density over clean states (utilising subsampling for computational feasibility), scoring each state via its negative log estimated density. It complements the Mahalanobis detector by avoiding assumptions regarding single-Gaussian shapes, though this flexibility comes at the expense of performance in higher dimensions.

\paragraph{Ensemble disagreement.}
Independently initialised predictors mapping $(s_t, a_t)$ to $s_{t+1}$ are trained on bootstrap samples derived from clean transitions \citep{chua2018,haider2023}. The resulting score is the mean population standard deviation across ensemble members for each state coordinate. This metric does not evaluate the predictions against the observed successor state.

\paragraph{Forward reconstruction error.}
This detector employs a single deterministic one-step forward-dynamics model to predict the next state from $(s_t,a_t)$, evaluating performance via the standardised residual between prediction and observation. It functions as the non-probabilistic, single-model equivalent to PEDM within the action-conditioned family.

\paragraph{State autoencoder.}
This detector uses an encoder--decoder network trained to reconstruct clean states, scoring each input by its reconstruction error. The underlying principle is that a model fitted exclusively to normal data will reproduce those inputs more accurately than anomalous ones; however, highly expressive autoencoders may occasionally reconstruct certain out-of-distribution samples with comparable fidelity.

\paragraph{CUSUM \citep{page1954}.}
This method standardises each state coordinate relative to its clean mean and spread, accumulating deviations over time in both directions using slack and decay parameters; the final score is derived from the larger accumulation across coordinates. Cumulative-sum charts are designed to detect sustained shifts following onset rather than isolated outliers.

\paragraph{Time-series features.}
This approach summarises a causal rolling window of recent observations by calculating per-coordinate descriptive statistics (means, spreads, and successive differences) and scoring the resulting feature vector against the clean distribution. It targets anomalies that disrupt local temporal structure rather than those affecting individual states.

\paragraph{Lyapunov divergence.}
This detector estimates local trajectory divergence—a finite-time Lyapunov-style statistic—within sliding windows to flag departures from the clean divergence profile. While it aims to detect shifts in local stability or phase-space geometry, these estimates are sensitive to noise and finite sample sizes.

\paragraph{Probabilistic Ensemble Dynamics Model (PEDM) \citep{haider2023,chua2018}.}
This method employs a bootstrapped ensemble of probabilistic forward-dynamics networks where each member samples candidate next states, scoring transitions by the minimum one-step prediction error among these samples. It represents the most robust published action-conditioned baseline in the collection.

\paragraph{DEXTER and DEXTER+C \citep{nasvytis2024}.}
The DEXTER approach extracts time-series features (tsfresh) for each state dimension independently over causal fixed-width windows, scoring each window with a dedicated isolation forest per dimension. DEXTER+C extends this by applying an episode-aware sequential CUSUM decision rule atop the base DEXTER scores.

\medskip
\noindent

All applicable detectors receive the same state representation, fitting trajectories, and independent calibration split. Alarm thresholds are established at the 95th percentile of clean calibration scores; for DEXTER+C, this value is calculated from the maximum cumulative statistic within each clean episode. Learned detectors are fitted using three random seeds, while analytic detectors plus DEXTER and DEXTER+C are fitted only once. The ACTI-Forward and reverse-only controls detailed in Section~\ref{sec:acti-variants} undergo evaluation under these same conditions, but they are mechanism ablations and therefore not included as suite members.

\subsection{Metrics and statistical analysis}
\label{sec:design-metrics}

The primary evaluation metrics utilise matched-time AUROC and AUPRC \citep{fawcett2006,davis2006}, which are inherently balanced because every post-onset anomalous score is paired with the clean twin's score at the same timestep, establishing a chance level of 0.5 for both metrics. We also report matched-time FPR at a 95\% TPR \citep{liang2018}, the alarm rate within clean episodes at the calibrated threshold, and detection delay (calculated as the first threshold crossing after onset, and right-censored at the episode horizon if a detector never fires).

Metrics are averaged for each combination of policy, anomaly, and severity tier, and statistical uncertainty is then evaluated across the policies, which serve as the units of replication. Each of the ten independently trained policies constitutes a full, independent repetition of the pipeline—including its own rollouts, detector fitting, and calibration—meaning confidence intervals and significance tests treat each policy as one observation within its environment. Episodes and timesteps within a single policy are not considered independent observations, as they share the policy's behaviour and the detectors fitted to its data; treating them as such would underestimate uncertainty. Macro summaries assign equal weight to tiers within anomalies, to anomalies within environments, and to environments collectively. We quantify uncertainty using a stratified policy-cluster bootstrap with 10,000 resamples to produce 95\% intervals.

The three pre-declared contrasts (ACTI minus STI, ACTI minus ACTI-Forward, and ACTI minus PEDM) undergo testing via paired policy-cluster sign-flip tests, with Holm correction \citep{holm1979} applied within each metric. Because the confidence interval estimation and hypothesis testing operate through different procedures, a bootstrap interval excluding zero may occasionally accompany a non-significant sign-flip $p$-value; in these cases, significance determinations follow the test results.

\section{Results}
\label{sec:results}

\subsection{Overall comparison}
\label{sec:results-main}
Table~\ref{tab:main-results} presents the matched-time AUROC metrics across the common support, detailing results per environment and as a macro average for all fifteen detectors. ACTI secures the top rank on the macro mean (\Result{main.matched_time_auroc.acti.macro}), trailed by its occupancy hybrid (\Result{main.matched_time_auroc.acti_occupancy.macro}) and state-only variant (\Result{main.matched_time_auroc.sti.macro}). Among baselines not based on irreversibility, PEDM delivers the best result (\Result{main.matched_time_auroc.pedm.macro}), surpassing forward reconstruction error (\Result{main.matched_time_auroc.reconstruction_error.macro}) and time-series features (\Result{main.matched_time_auroc.time_series_features.macro}). DEXTER (\Result{main.matched_time_auroc.dexter.macro}) and DEXTER+C (\Result{main.matched_time_auroc.dexter_c.macro}) occupy the space between forward reconstruction error and ensemble disagreement within this hierarchy. The weakest outcomes for these window-based detectors occur in Acrobot (\Result{point.matched_time_auroc.dexter.ac}) and Breakout (\Result{point.matched_time_auroc.dexter.br} for DEXTER), corresponding to environments where margins for the transition-irreversibility family are greatest.  Classical marginal-state detectors lag behind (Mahalanobis \Result{main.matched_time_auroc.mahalanobis_distance.macro}, KDE \Result{main.matched_time_auroc.density_kde.macro}), while the nonlinear-dynamics detector records the poorest performance (\Result{main.matched_time_auroc.lyapunov_divergence.macro}), falling below chance in the visual environment. The corresponding matched-time AUPRC results are reported in Table~\ref{tab:auprc-results} in Appendix~\ref{app:auprc-results}.

\begin{table}[!htbp]
  \centering
  \footnotesize
  \setlength{\tabcolsep}{2pt}
  \begin{tabular*}{\textwidth}{@{\extracolsep{\fill}}lcccc@{}}
    \toprule
    Detector & LunarLander & Acrobot & HalfCheetah & Ant \\
    \midrule
    Mahalanobis distance & \ResultCell{main.matched_time_auroc.mahalanobis_distance.ll} & \ResultCell{main.matched_time_auroc.mahalanobis_distance.ac} & \ResultCell{main.matched_time_auroc.mahalanobis_distance.hc} & \ResultCell{main.matched_time_auroc.mahalanobis_distance.an} \\
    KDE density & \ResultCell{main.matched_time_auroc.density_kde.ll} & \ResultCell{main.matched_time_auroc.density_kde.ac} & \ResultCell{main.matched_time_auroc.density_kde.hc} & \ResultCell{main.matched_time_auroc.density_kde.an} \\
    Ensemble disagreement & \ResultCell{main.matched_time_auroc.ensemble_disagreement.ll} & \ResultCell{main.matched_time_auroc.ensemble_disagreement.ac} & \ResultCell{main.matched_time_auroc.ensemble_disagreement.hc} & \ResultCell{main.matched_time_auroc.ensemble_disagreement.an} \\
    Forward reconstruction error & \ResultCell{main.matched_time_auroc.reconstruction_error.ll} & \ResultCell{main.matched_time_auroc.reconstruction_error.ac} & \ResultCell{main.matched_time_auroc.reconstruction_error.hc} & \ResultCell{main.matched_time_auroc.reconstruction_error.an} \\
    State autoencoder & \ResultCell{main.matched_time_auroc.autoencoder.ll} & \ResultCell{main.matched_time_auroc.autoencoder.ac} & \ResultCell{main.matched_time_auroc.autoencoder.hc} & \ResultCell{main.matched_time_auroc.autoencoder.an} \\
    CUSUM & \ResultCell{main.matched_time_auroc.cusum.ll} & \ResultCell{main.matched_time_auroc.cusum.ac} & \ResultCell{main.matched_time_auroc.cusum.hc} & \ResultCell{main.matched_time_auroc.cusum.an} \\
    Time-series features & \ResultCell{main.matched_time_auroc.time_series_features.ll} & \ResultCell{main.matched_time_auroc.time_series_features.ac} & \ResultCell{main.matched_time_auroc.time_series_features.hc} & \ResultCell{main.matched_time_auroc.time_series_features.an} \\
    DEXTER & \ResultCell{main.matched_time_auroc.dexter.ll} & \ResultCell{main.matched_time_auroc.dexter.ac} & \ResultCell{main.matched_time_auroc.dexter.hc} & \ResultCell{main.matched_time_auroc.dexter.an} \\
    DEXTER+C & \ResultCell{main.matched_time_auroc.dexter_c.ll} & \ResultCell{main.matched_time_auroc.dexter_c.ac} & \ResultCell{main.matched_time_auroc.dexter_c.hc} & \ResultCell{main.matched_time_auroc.dexter_c.an} \\
    Lyapunov divergence & \ResultCell{main.matched_time_auroc.lyapunov_divergence.ll} & \ResultCell{main.matched_time_auroc.lyapunov_divergence.ac} & \ResultCell{main.matched_time_auroc.lyapunov_divergence.hc} & \ResultCell{main.matched_time_auroc.lyapunov_divergence.an} \\
    PEDM & \ResultCell{main.matched_time_auroc.pedm.ll} & \ResultCell{main.matched_time_auroc.pedm.ac} & \ResultCell{main.matched_time_auroc.pedm.hc} & \ResultCell{main.matched_time_auroc.pedm.an} \\
    STI & \ResultCell{main.matched_time_auroc.sti.ll} & \ResultCell{main.matched_time_auroc.sti.ac} & \ResultCell{main.matched_time_auroc.sti.hc} & \ResultCell{main.matched_time_auroc.sti.an} \\
    ACTI & \ResultCellBest{main.matched_time_auroc.acti.ll} & \ResultCell{main.matched_time_auroc.acti.ac} & \ResultCellBest{main.matched_time_auroc.acti.hc} & \ResultCellBest{main.matched_time_auroc.acti.an} \\
    Occupancy-only & \ResultCell{main.matched_time_auroc.occupancy_only.ll} & \ResultCell{main.matched_time_auroc.occupancy_only.ac} & \ResultCell{main.matched_time_auroc.occupancy_only.hc} & \ResultCell{main.matched_time_auroc.occupancy_only.an} \\
    ACTI+O & \ResultCell{main.matched_time_auroc.acti_occupancy.ll} & \ResultCellBest{main.matched_time_auroc.acti_occupancy.ac} & \ResultCell{main.matched_time_auroc.acti_occupancy.hc} & \ResultCell{main.matched_time_auroc.acti_occupancy.an} \\
    \bottomrule
  \end{tabular*}
  \par\medskip
  \begin{tabular*}{\textwidth}{@{\extracolsep{\fill}}lccc@{}}
    \toprule
    Detector & Humanoid & Breakout & Macro mean \\
    \midrule
    Mahalanobis distance & \ResultCell{main.matched_time_auroc.mahalanobis_distance.hu} & \ResultCell{main.matched_time_auroc.mahalanobis_distance.br} & \ResultCell{main.matched_time_auroc.mahalanobis_distance.macro} \\
    KDE density & \ResultCell{main.matched_time_auroc.density_kde.hu} & \ResultCell{main.matched_time_auroc.density_kde.br} & \ResultCell{main.matched_time_auroc.density_kde.macro} \\
    Ensemble disagreement & \ResultCell{main.matched_time_auroc.ensemble_disagreement.hu} & \ResultCell{main.matched_time_auroc.ensemble_disagreement.br} & \ResultCell{main.matched_time_auroc.ensemble_disagreement.macro} \\
    Forward reconstruction error & \ResultCell{main.matched_time_auroc.reconstruction_error.hu} & \ResultCell{main.matched_time_auroc.reconstruction_error.br} & \ResultCell{main.matched_time_auroc.reconstruction_error.macro} \\
    State autoencoder & \ResultCell{main.matched_time_auroc.autoencoder.hu} & \ResultCell{main.matched_time_auroc.autoencoder.br} & \ResultCell{main.matched_time_auroc.autoencoder.macro} \\
    CUSUM & \ResultCell{main.matched_time_auroc.cusum.hu} & \ResultCell{main.matched_time_auroc.cusum.br} & \ResultCell{main.matched_time_auroc.cusum.macro} \\
    Time-series features & \ResultCell{main.matched_time_auroc.time_series_features.hu} & \ResultCell{main.matched_time_auroc.time_series_features.br} & \ResultCell{main.matched_time_auroc.time_series_features.macro} \\
    DEXTER & \ResultCell{main.matched_time_auroc.dexter.hu} & \ResultCell{main.matched_time_auroc.dexter.br} & \ResultCell{main.matched_time_auroc.dexter.macro} \\
    DEXTER+C & \ResultCell{main.matched_time_auroc.dexter_c.hu} & \ResultCell{main.matched_time_auroc.dexter_c.br} & \ResultCell{main.matched_time_auroc.dexter_c.macro} \\
    Lyapunov divergence & \ResultCell{main.matched_time_auroc.lyapunov_divergence.hu} & \ResultCell{main.matched_time_auroc.lyapunov_divergence.br} & \ResultCell{main.matched_time_auroc.lyapunov_divergence.macro} \\
    PEDM & \ResultCell{main.matched_time_auroc.pedm.hu} & \ResultCell{main.matched_time_auroc.pedm.br} & \ResultCell{main.matched_time_auroc.pedm.macro} \\
    STI & \ResultCell{main.matched_time_auroc.sti.hu} & \ResultCell{main.matched_time_auroc.sti.br} & \ResultCell{main.matched_time_auroc.sti.macro} \\
    ACTI & \ResultCellBest{main.matched_time_auroc.acti.hu} & \ResultCellBest{main.matched_time_auroc.acti.br} & \ResultCellBest{main.matched_time_auroc.acti.macro} \\
    Occupancy-only & \ResultCell{main.matched_time_auroc.occupancy_only.hu} & \ResultCell{main.matched_time_auroc.occupancy_only.br} & \ResultCell{main.matched_time_auroc.occupancy_only.macro} \\
    ACTI+O & \ResultCell{main.matched_time_auroc.acti_occupancy.hu} & \ResultCell{main.matched_time_auroc.acti_occupancy.br} & \ResultCell{main.matched_time_auroc.acti_occupancy.macro} \\
    \bottomrule
  \end{tabular*}
\normalsize
  \caption{Policy-level matched-time AUROC on the frozen common support $A^*$ (four mechanisms, five tiers, tier-averaged), mean and 95\% policy-bootstrap interval. Macro weights environments equally. Bold marks the highest value in each column.}
  \label{tab:main-results}
\end{table}

\begin{figure}[!htbp]
  \centering
  \includegraphics[width=\linewidth]{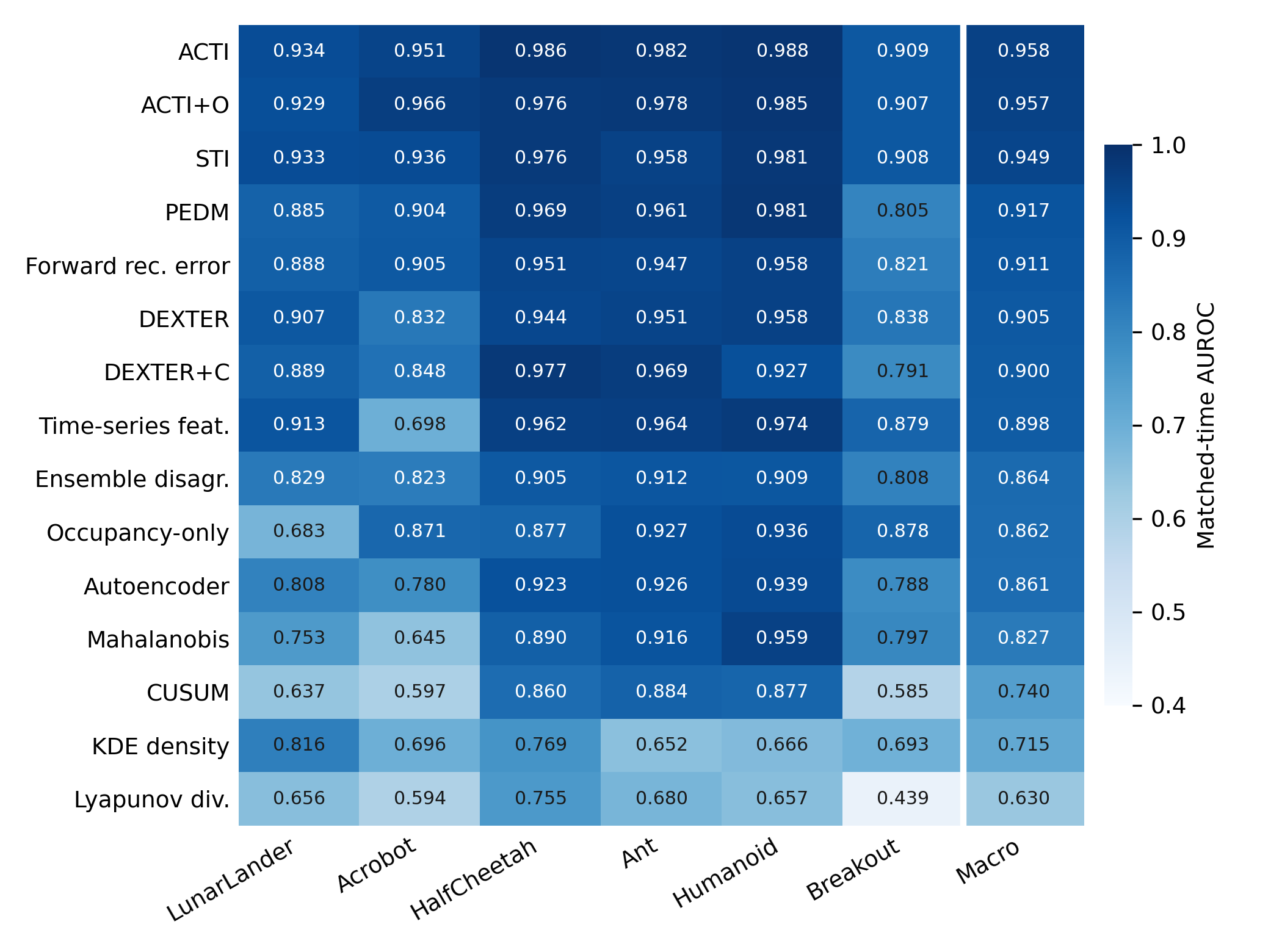}
  \caption{Environment-level matched-time AUROC on the frozen common
  support. Cells average equally across qualified anomaly tiers at the
  policy level.}
  \label{fig:matched-time-heatmap}
\end{figure}

Across the environments (Figure~\ref{fig:matched-time-heatmap}), ACTI holds the highest point estimate in five of the six scenarios; on Acrobot the occupancy hybrid leads and ACTI ranks second. Sequential DEXTER+C records its strongest results on HalfCheetah (\Result{point.matched_time_auroc.dexter_c.hc_an}) and Ant, where it is the closest competitor outside the transition-irreversibility family, with intervals overlapping those of the leaders. The most substantial performance margins of the transition-irreversibility family over PEDM occur on Breakout (\Result{main.matched_time_auroc.acti.br} versus \Result{main.matched_time_auroc.pedm.br}); on the MuJoCo tasks the corresponding margins range from $0.007$ (Humanoid) to $0.021$ (Ant). While no detector claims the top score across all instances, the transition-irreversibility family nonetheless secures the top three macro positions.

\subsection{Pre-declared contrasts}
\label{sec:results-contrasts}

Table~\ref{tab:contrast-results} reports the three pre-declared paired contrasts across four metrics, with two of the comparisons favouring ACTI.

Against PEDM, ACTI increases AUROC by \Result{contrast.matched_time_auroc.acti_minus_pedm} ($p_{\mathrm{Holm}}=\Result{confirmatory.matched_time_auroc.acti_minus_pedm.holm_adjusted_p}$), boosts AUPRC by \Result{contrast.matched_time_auprc.acti_minus_pedm} ($p=\Result{confirmatory.matched_time_auprc.acti_minus_pedm.holm_adjusted_p}$), and has a lower FPR at 95\% TPR (ACTI minus PEDM, \Result{contrast.matched_time_fpr_at_95tpr.acti_minus_pedm}, $p=\Result{confirmatory.matched_time_fpr_at_95tpr.acti_minus_pedm.holm_adjusted_p}$); the detection-delay difference is not significant (\Result{contrast.detection_delay.acti_minus_pedm}). 

\begin{figure}[!htbp]
  \centering
  \includegraphics[width=0.82\linewidth]{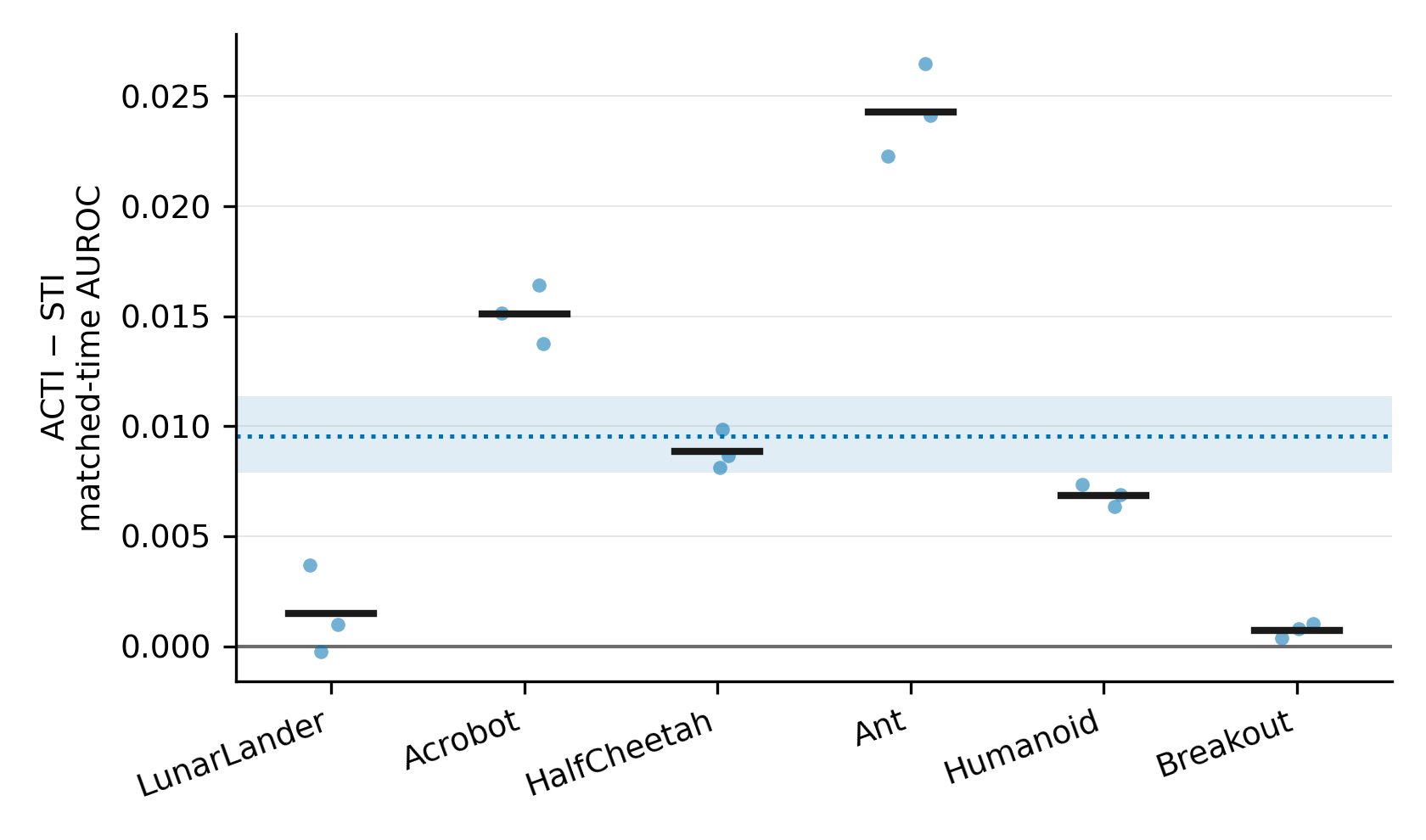}
  \caption{Paired ACTI-minus-STI matched-time AUROC by environment, with the stratified macro policy-bootstrap interval. The action-conditioning gain is small but consistently positive.}
  \label{fig:acti-minus-sti}
\end{figure}

Against STI, action conditioning delivers small but consistent discrimination gains (AUROC \Result{contrast.matched_time_auroc.acti_minus_sti}, $p=\Result{confirmatory.matched_time_auroc.acti_minus_sti.holm_adjusted_p}$; AUPRC \Result{contrast.matched_time_auprc.acti_minus_sti}, $p=\Result{confirmatory.matched_time_auprc.acti_minus_sti.holm_adjusted_p}$; FPR \Result{contrast.matched_time_fpr_at_95tpr.acti_minus_sti}, $p=\Result{confirmatory.matched_time_fpr_at_95tpr.acti_minus_sti.holm_adjusted_p}$) and a shorter detection delay (ACTI minus STI, \Result{contrast.detection_delay.acti_minus_sti} steps, $p=\Result{confirmatory.detection_delay.acti_minus_sti.holm_adjusted_p}$). Figure~\ref{fig:acti-minus-sti} displays the paired per-environment AUROC differences. In these experiments, conditioning both the forward and inverse models on the recorded commanded action enhances discriminative performance and accelerates detection relative to the state-only variant.

\begin{table}[!htbp]
  \centering
  \footnotesize
  \setlength{\tabcolsep}{3pt}%
  \begin{tabularx}{\textwidth}{@{}>{\raggedright\arraybackslash}p{.23\textwidth}YYY@{}}
    \toprule
    Metric & ACTI $-$ STI & ACTI $-$ ACTI-Forward & ACTI $-$ PEDM \\
    \midrule
    Matched-time AUROC & \Result{contrast.matched_time_auroc.acti_minus_sti} ($p=\Result{confirmatory.matched_time_auroc.acti_minus_sti.holm_adjusted_p}$) & \Result{contrast.matched_time_auroc.acti_minus_forward_only} ($p=\Result{confirmatory.matched_time_auroc.acti_minus_forward_only.holm_adjusted_p}$) & \Result{contrast.matched_time_auroc.acti_minus_pedm} ($p=\Result{confirmatory.matched_time_auroc.acti_minus_pedm.holm_adjusted_p}$) \\
    Matched-time AUPRC & \Result{contrast.matched_time_auprc.acti_minus_sti} ($p=\Result{confirmatory.matched_time_auprc.acti_minus_sti.holm_adjusted_p}$) & \Result{contrast.matched_time_auprc.acti_minus_forward_only} ($p=\Result{confirmatory.matched_time_auprc.acti_minus_forward_only.holm_adjusted_p}$) & \Result{contrast.matched_time_auprc.acti_minus_pedm} ($p=\Result{confirmatory.matched_time_auprc.acti_minus_pedm.holm_adjusted_p}$) \\
    Matched-time FPR@95\%TPR & \Result{contrast.matched_time_fpr_at_95tpr.acti_minus_sti} ($p=\Result{confirmatory.matched_time_fpr_at_95tpr.acti_minus_sti.holm_adjusted_p}$) & \Result{contrast.matched_time_fpr_at_95tpr.acti_minus_forward_only} ($p=\Result{confirmatory.matched_time_fpr_at_95tpr.acti_minus_forward_only.holm_adjusted_p}$) & \Result{contrast.matched_time_fpr_at_95tpr.acti_minus_pedm} ($p=\Result{confirmatory.matched_time_fpr_at_95tpr.acti_minus_pedm.holm_adjusted_p}$) \\
    Detection delay (steps) & \Result{contrast.detection_delay.acti_minus_sti} ($p=\Result{confirmatory.detection_delay.acti_minus_sti.holm_adjusted_p}$) & \Result{contrast.detection_delay.acti_minus_forward_only} ($p=\Result{confirmatory.detection_delay.acti_minus_forward_only.holm_adjusted_p}$) & \Result{contrast.detection_delay.acti_minus_pedm} ($p=\Result{confirmatory.detection_delay.acti_minus_pedm.holm_adjusted_p}$) \\
    \bottomrule
  \end{tabularx}
\normalsize
  \caption{Pre-declared paired policy-level contrasts (ACTI minus the named baseline) with 95\% bootstrap intervals; $p$-values are Holm-adjusted policy-cluster sign-flip tests within each metric. Negative delay and FPR differences favour ACTI.}
  \label{tab:contrast-results}
\end{table}

\begin{figure}[!htbp]
  \centering
  \includegraphics[width=0.82\linewidth]{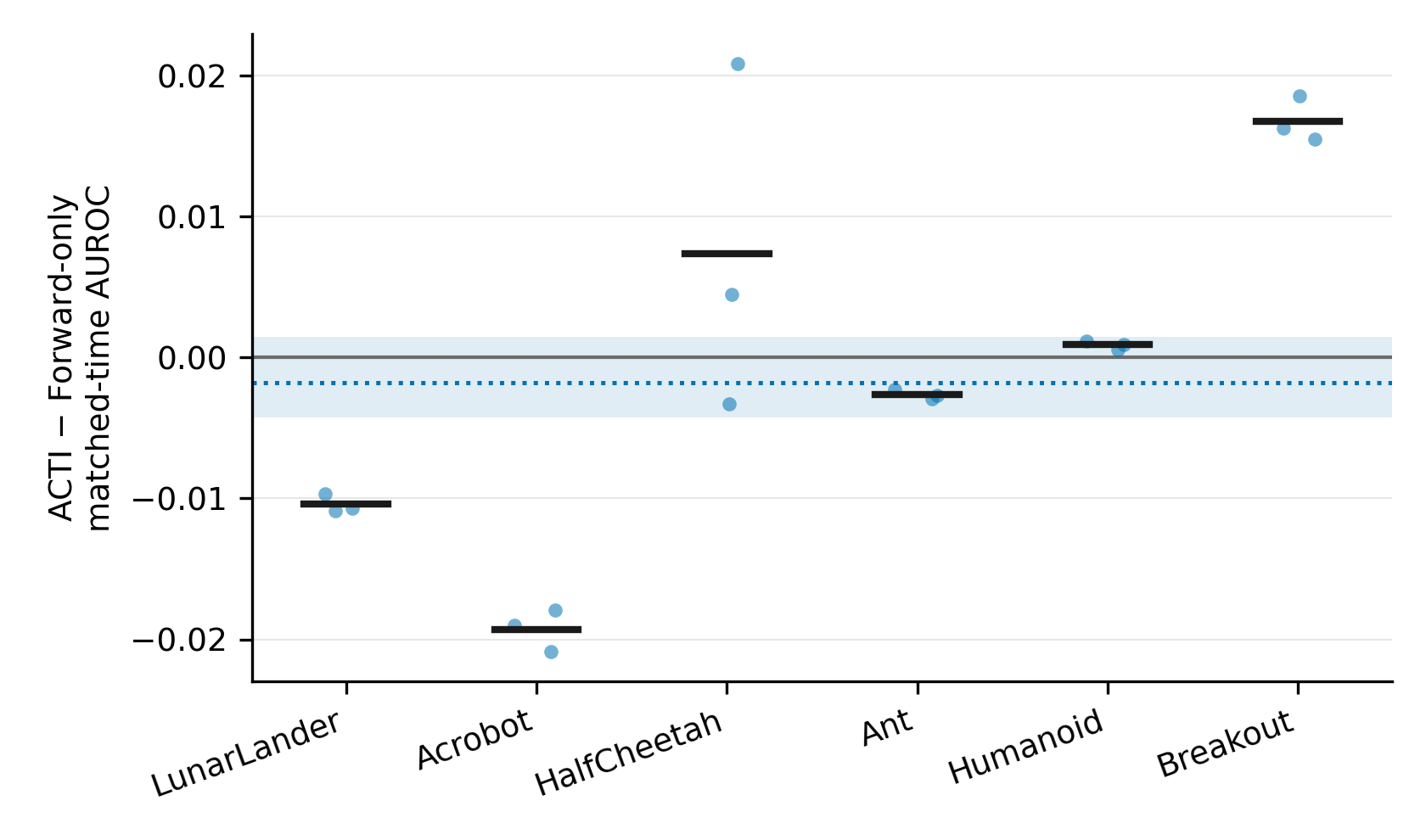}
  \caption{Paired ACTI-minus-ACTI-Forward matched-time AUROC by environment. The aggregate 95\% interval includes zero, indicating that the Holm-adjusted test detects no statistically significant disparity in AUROC. Note that this result does not establish equivalence.}
  \label{fig:acti-minus-forward}
\end{figure}

The third pre-declared comparison assesses whether the forward--inverse likelihood contrast enhances detection relative to the capacity-matched ACTI-Forward baseline. Values presented here are ACTI minus ACTI-Forward. Holm-adjusted analysis revealed no statistically significant divergence in matched-time AUROC (\Result{contrast.matched_time_auroc.acti_minus_forward_only}, $p=\Result{confirmatory.matched_time_auroc.acti_minus_forward_only.holm_adjusted_p}$) or FPR at 95\% TPR (\Result{contrast.matched_time_fpr_at_95tpr.acti_minus_forward_only}, $p=\Result{confirmatory.matched_time_fpr_at_95tpr.acti_minus_forward_only.holm_adjusted_p}$). Conversely, ACTI-Forward attained significantly better matched-time AUPRC (\Result{contrast.matched_time_auprc.acti_minus_forward_only}, $p=\Result{confirmatory.matched_time_auprc.acti_minus_forward_only.holm_adjusted_p}$) and exhibited a shorter detection delay (\Result{contrast.detection_delay.acti_minus_forward_only} steps, $p=\Result{confirmatory.detection_delay.acti_minus_forward_only.holm_adjusted_p}$). Figure~\ref{fig:acti-minus-forward} shows the paired AUROC discrepancies across individual environments. With respect to the reverse-only baseline, the disparity between ACTI and this control yielded values of \Result{contrast.matched_time_auroc.acti_minus_reverse_only} for AUROC and \Result{contrast.detection_delay.acti_minus_reverse_only} steps for detection delay; in both instances, the reported 95\% intervals include zero. These findings do not establish that the detectors are statistically equivalent. Section~\ref{sec:discussion} considers potential causes for the absence of a noticeable benefit attributable to the contrast.

\subsection{Component analysis}
\label{sec:results-components}

Table~\ref{tab:ablation-results} reports the effects of the tested component changes. At severity tier $0.6$, increasing the smoothing window from $L=1$ to $L=20$ raises matched-time AUROC by \Result{contrast.matched_time_auroc.window_20_minus_window_1}, incurring a latency penalty of \Result{contrast.detection_delay.window_20_minus_window_1} steps. Per-timestep clean-profile calibration enhances both metrics relative to global calibration (\Result{contrast.matched_time_auroc.timestep_mean_sd_minus_global_mean_sd} in AUROC and \Result{contrast.detection_delay.timestep_mean_sd_minus_global_mean_sd} steps of delay); the drift within episode scores is substantial enough that correcting it aids detection, even though matched-time evaluation negates its direct effect on the metric. Across all five severity tiers, ACTI improves AUROC relative to STI by \Result{contrast.matched_time_auroc.acti_minus_sti} and reduces detection delay by approximately \Result{point.contrast_delay.acti_minus_sti_rounded} steps (ACTI minus STI: \Result{contrast.detection_delay.acti_minus_sti} steps). Conversely, the occupancy hybrid yields no advantage (ACTI+O minus ACTI: \Result{contrast.matched_time_auroc.acti_occupancy_minus_acti} in AUROC and \Result{contrast.detection_delay.acti_occupancy_minus_acti} steps of delay), while standalone occupancy ranks tenth.

\begin{table}[!htbp]
  \centering
  \small
  \begin{tabularx}{\textwidth}{@{}YccY@{}}
    \toprule
    Contrast & AUROC effect & Delay effect & Interpretation \\
    \midrule
    ACTI $-$ ACTI-Forward & \Result{contrast.matched_time_auroc.acti_minus_forward_only} & \Result{contrast.detection_delay.acti_minus_forward_only} & No significant AUROC difference; ACTI-Forward is faster \\
    ACTI $-$ reverse-only & \Result{contrast.matched_time_auroc.acti_minus_reverse_only} & \Result{contrast.detection_delay.acti_minus_reverse_only} & AUROC and delay intervals include zero \\
    ACTI $-$ STI & \Result{contrast.matched_time_auroc.acti_minus_sti} & \Result{contrast.detection_delay.acti_minus_sti} & Action conditioning: accuracy and latency \\
    ACTI+O $-$ ACTI & \Result{contrast.matched_time_auroc.acti_occupancy_minus_acti} & \Result{contrast.detection_delay.acti_occupancy_minus_acti} & Occupancy term: no gain, slower \\
    Timestep $-$ global calibration & \Result{contrast.matched_time_auroc.timestep_mean_sd_minus_global_mean_sd} & \Result{contrast.detection_delay.timestep_mean_sd_minus_global_mean_sd} & Drift correction: both axes improve \\
    Window 20 $-$ window 1 & \Result{contrast.matched_time_auroc.window_20_minus_window_1} & \Result{contrast.detection_delay.window_20_minus_window_1} & Higher AUROC; longer delay \\
    \bottomrule
  \end{tabularx}
\normalsize
  \caption{Paired macro component effects on matched-time AUROC and detection delay, each with a 95\% interval. ACTI minus ACTI-Forward, ACTI minus reverse-only, ACTI minus STI, and ACTI+O minus ACTI average all five severity tiers. Calibration and smoothing-window comparisons use the middle severity tier.}
  \label{tab:ablation-results}
\end{table}

\begin{figure}[!htbp]
  \centering
  \includegraphics[width=0.82\linewidth]{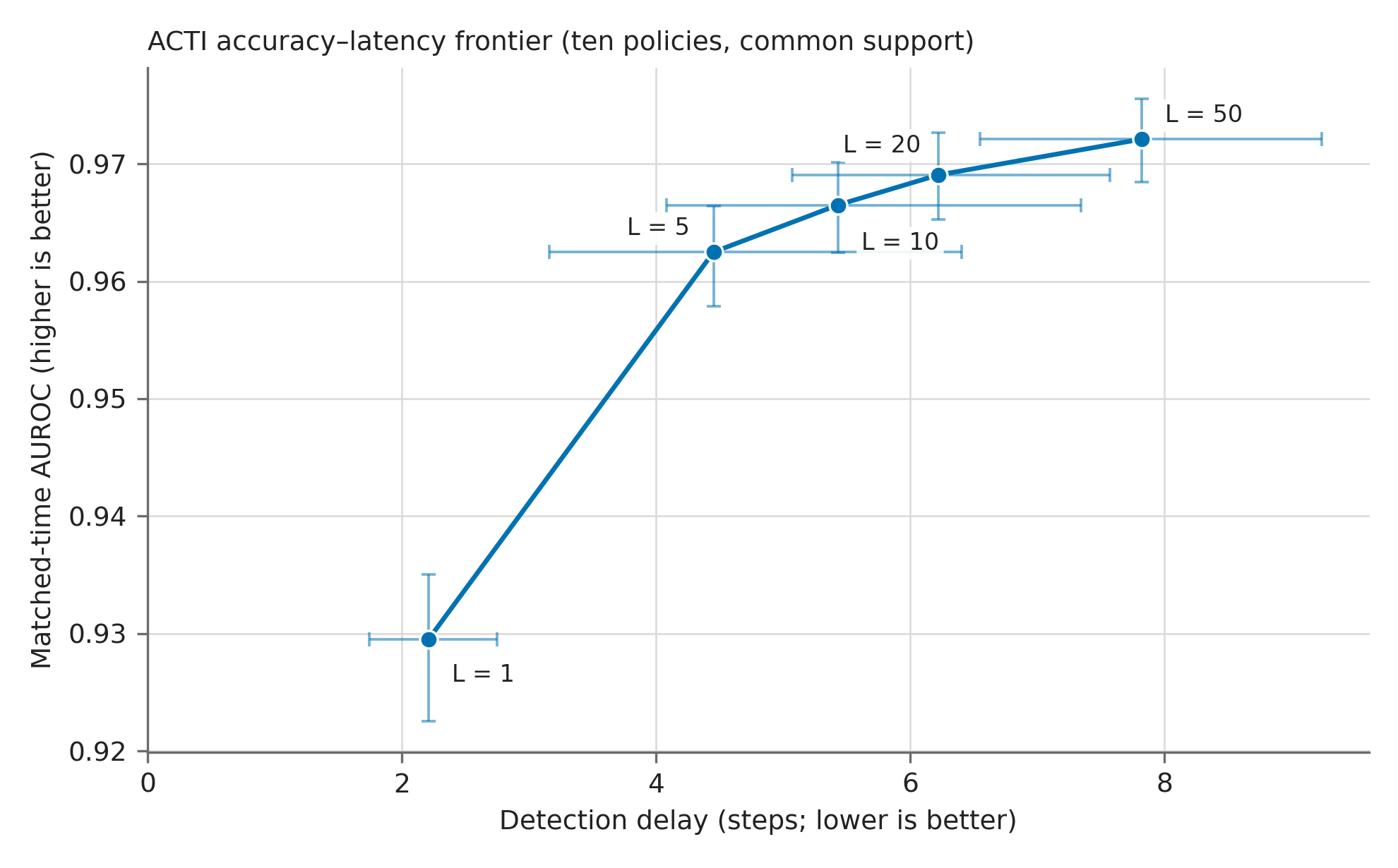}
  \caption{Accuracy--latency frontier over causal smoothing windows $L\in\{1,5,10,20,50\}$ on common support at severity tier $0.6$. Across six environments, estimates use ten policies per environment, with equal environment weights and 95\% policy-bootstrap intervals.}
  \label{fig:accuracy-latency}

\end{figure}

Figure~\ref{fig:accuracy-latency} reports the accuracy--latency frontier on common support at severity tier $0.6$, using ten policies per environment. Increasing the smoothing window from $L=1$ to $L=20$ raises macro matched-time AUROC from \Result{frontier.matched_time_auroc.l1} to \Result{frontier.matched_time_auroc.l20}, while detection delay increases from \Result{frontier.detection_delay.l1} to \Result{frontier.detection_delay.l20} steps. The window-size choice is therefore dependent on the acceptable accuracy--latency trade-off.

\subsection{Severity response, environments, and expanded support}
\label{sec:results-environments}

\begin{figure}[!htbp]
  \centering
  \includegraphics[width=0.86\linewidth]{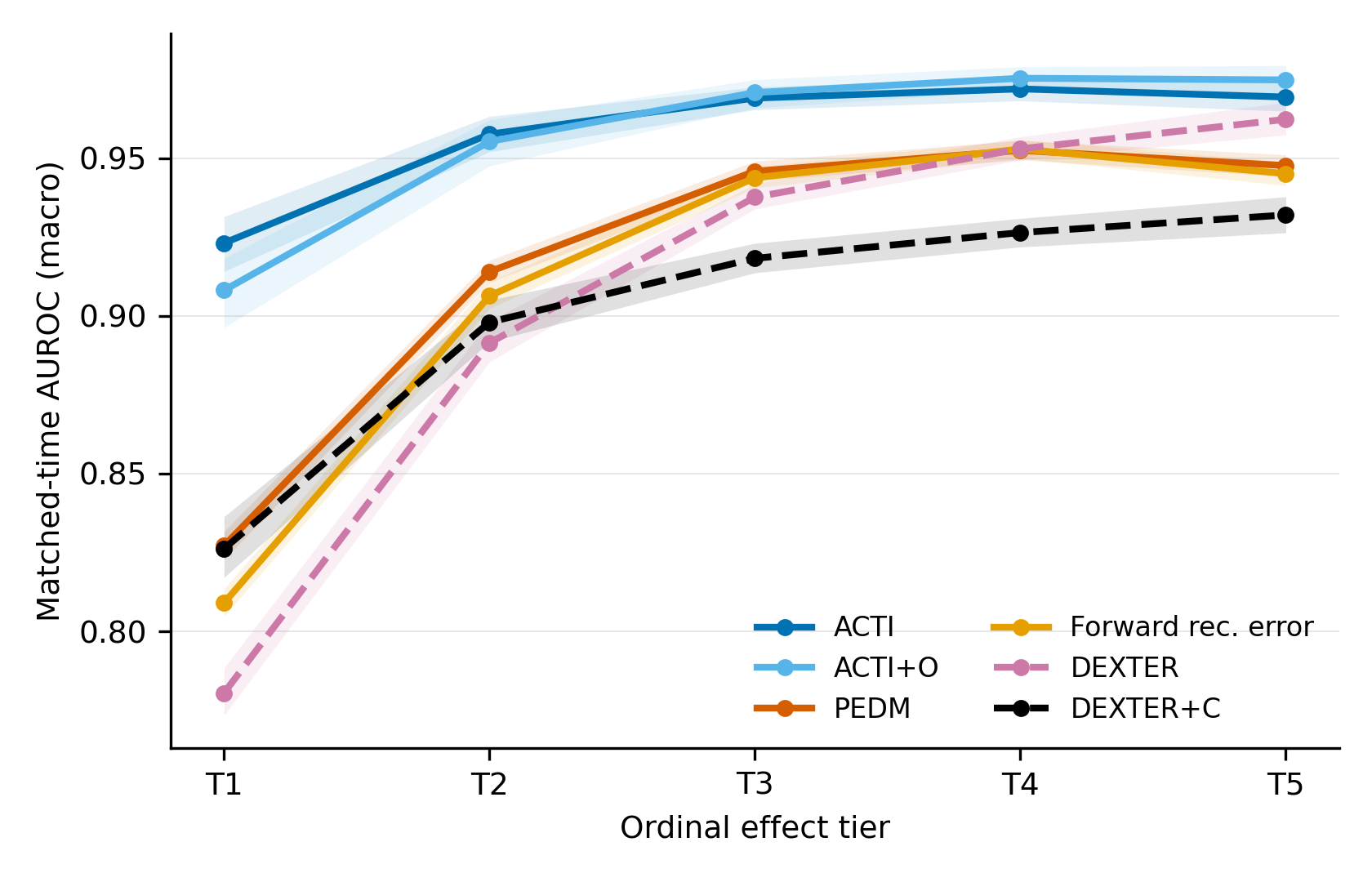}
  \caption{Matched-time AUROC across the five ordinal effect tiers for ACTI and ACTI+O, representative transition baselines, and the window-based DEXTER variants (dashed) on common support. Shading is the 95\% policy-bootstrap interval; the advantage of ACTI and ACTI+O is concentrated at low tiers.}
  \label{fig:severity-curves}
\end{figure}

Figure~\ref{fig:severity-curves} displays the macro matched-time AUROC across five ordinal effect tiers. ACTI leads at the two weakest tiers, and ACTI+O is marginally higher from the middle tier upward; the family's performance margin peaks where detection proves most challenging. At the weakest tier, ACTI achieves \Result{severity.tier1.matched_time_auroc.acti} compared to \Result{severity.tier1.matched_time_auroc.pedm} for PEDM, \Result{severity.tier1.matched_time_auroc.reconstruction_error} for forward reconstruction error, and \Result{severity.tier1.matched_time_auroc.dexter} for DEXTER; this gap narrows to $\approx 0.01$--$0.02$ at the upper tiers (reaching \Result{severity.tier5.matched_time_auroc.dexter} for DEXTER). These findings indicate that the calibrated, smoothed transition statistic is particularly effective for identifying subtle anomalies, as all transition-model detectors perform comparably across the highest tiers.

The fifteen environment-specific qualified cells in expanded support do not retain a uniform detector ordering (Table~\ref{tab:expanded-auroc-results} in Appendix~\ref{app:expanded-auroc-results}). ACTI attains \Result{expanded.matched_time_auroc.acti.correlation_break_br_hu} on both Breakout and Humanoid (correlation break), alongside \Result{expanded.matched_time_auroc.acti.dynamics_scaling_ac} on Acrobot dynamics scaling. Over all cells, DEXTER ranges from \Result{expanded.matched_time_auroc.dexter.min} to \Result{expanded.matched_time_auroc.dexter.max}, and DEXTER+C from \Result{expanded.matched_time_auroc.dexter_c.min} to \Result{expanded.matched_time_auroc.dexter_c.max}. Observation delay proves challenging, with ACTI achieving a matched-time AUROC of approximately \Result{expanded.matched_time_auroc.acti.observation_delay_best} on LunarLander and Breakout. Delayed observations remain situated on the in-distribution manifold, requiring temporally or action-conditioned context to expose them. Furthermore, while the transition-irreversibility family displays a distinct performance advantage over PEDM on Breakout's action-conditioned dynamics cell (STI/ACTI $\approx$\Result{expanded.matched_time_auroc.acti_sti.action_conditioned_br} versus PEDM at \Result{expanded.matched_time_auroc.pedm.action_conditioned_br}), every detector struggles on Acrobot's corresponding cell ($\le$\Result{expanded.matched_time_auroc.action_conditioned_ac.max}). Overall, the transition-irreversibility family achieves the highest mean matched-time AUROC in ten of the fifteen expanded-support cells, and is in the top three in every cell. PEDM, on the other hand, ranks first in only two cells and is in the top three in five out of fifteen.

\begin{figure}[!htbp]
  \centering
  \includegraphics[width=0.78\linewidth]{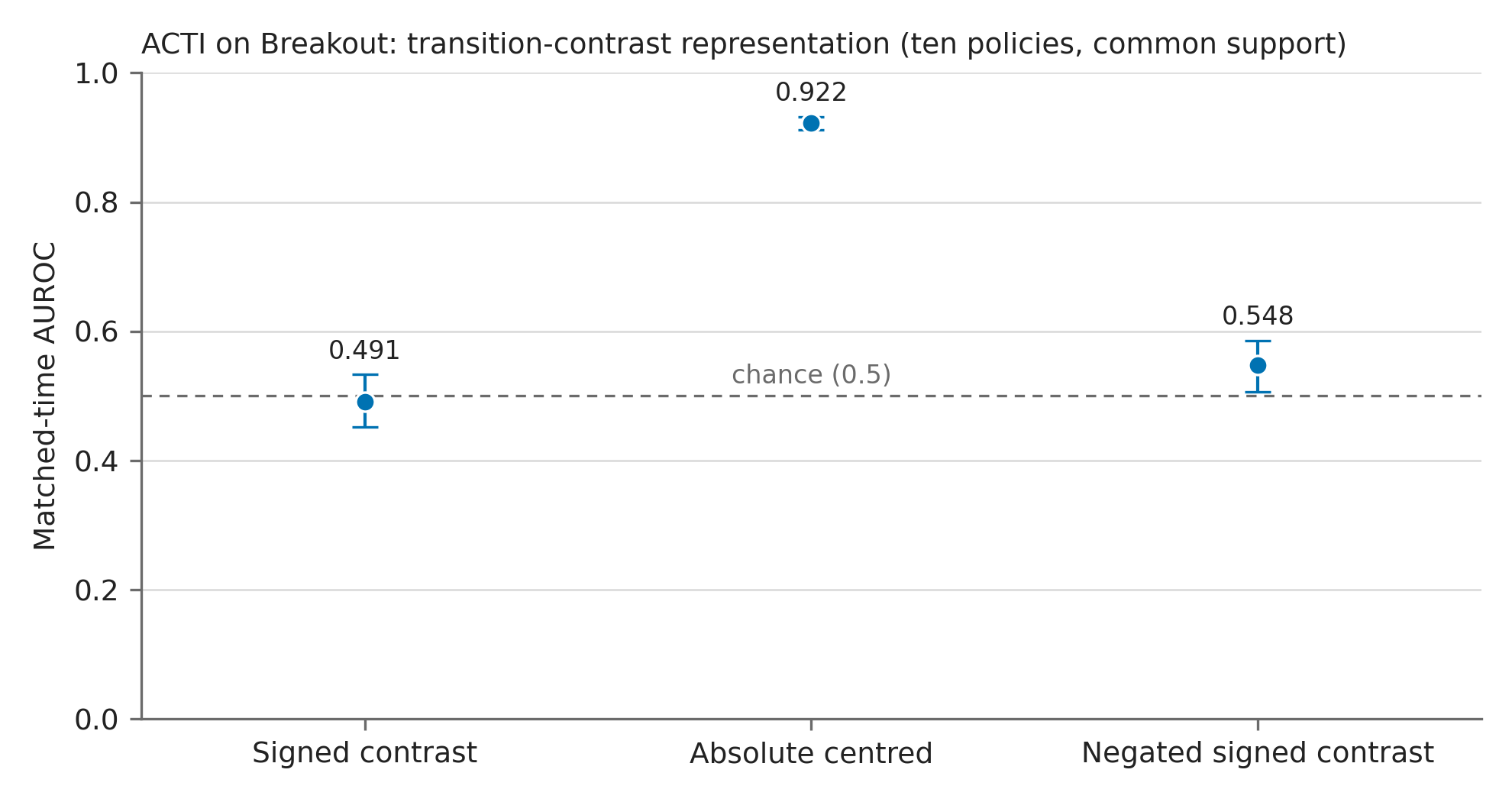}
  \caption{Breakout contrast-to-score transformation ablation on common support at the middle severity tier. The state representation and fitted forward and reverse models remain fixed; only the transformation of their log-likelihood contrast changes. Points and 95\% intervals aggregate across the ten policy seeds.}
  \label{fig:breakout-representation}
\end{figure}

Figure~\ref{fig:breakout-representation} examines the conversion of the forward–reverse likelihood contrast into a detection score for Breakout on common support. It contrasts the signed variant, its negation, and the absolute deviation from the clean per-timestep profile, with the state representation and fitted models held fixed. The absolute centred score achieves a matched-time AUROC of 0.922, compared with 0.491 and 0.548 for the signed and negated forward-reverse log-likelihood contrasts, respectively. These results support using deviations from the normal contrast in either direction as the anomaly signal.

\subsection{Operating points and computational cost}
\label{sec:results-compute}

Table~\ref{tab:operating-points} details the three operating-point metrics. Clean-timestep alarm rates are \Result{point.clean_alarm_rate.suite} across all detectors, with the exception of DEXTER+C (\Result{point.clean_alarm_rate.dexter_c.macro}), whose sequential threshold is calibrated using clean episode maxima. The similar clean alarm rates indicate comparable calibration for the remaining detectors. The matched-time FPR at 95\% TPR creates a distinct separation within the suite: ACTI and ACTI+O lead (\Result{operating.matched_time_fpr_at_95tpr.acti.macro} and \Result{operating.matched_time_fpr_at_95tpr.acti_occupancy.macro}), followed by STI (\Result{operating.matched_time_fpr_at_95tpr.sti.macro}), DEXTER (\Result{operating.matched_time_fpr_at_95tpr.dexter.macro}), and PEDM (\Result{operating.matched_time_fpr_at_95tpr.pedm.macro}); the marginal-state and sequential detectors suffer from elevated false-positive rates at 95\% recall.

\begin{table}[!htbp]
  \centering
  \small
  \begin{tabularx}{\textwidth}{@{}Yccc@{}}
    \toprule
    Detector & \shortstack{Clean alarm rate} & Matched-time FPR at 95\% TPR & Detection delay (steps) \\
    \midrule
    Mahalanobis distance & \ResultCell{operating.clean_alarm_rate.mahalanobis_distance.macro} & \ResultCell{operating.matched_time_fpr_at_95tpr.mahalanobis_distance.macro} & \ResultCell{operating.detection_delay.mahalanobis_distance.macro} \\
    KDE density & \ResultCell{operating.clean_alarm_rate.density_kde.macro} & \ResultCell{operating.matched_time_fpr_at_95tpr.density_kde.macro} & \ResultCell{operating.detection_delay.density_kde.macro} \\
    Ensemble disagreement & \ResultCell{operating.clean_alarm_rate.ensemble_disagreement.macro} & \ResultCell{operating.matched_time_fpr_at_95tpr.ensemble_disagreement.macro} & \ResultCell{operating.detection_delay.ensemble_disagreement.macro} \\
    Forward reconstruction error & \ResultCell{operating.clean_alarm_rate.reconstruction_error.macro} & \ResultCell{operating.matched_time_fpr_at_95tpr.reconstruction_error.macro} & \ResultCell{operating.detection_delay.reconstruction_error.macro} \\
    State autoencoder & \ResultCell{operating.clean_alarm_rate.autoencoder.macro} & \ResultCell{operating.matched_time_fpr_at_95tpr.autoencoder.macro} & \ResultCell{operating.detection_delay.autoencoder.macro} \\
    CUSUM & \ResultCell{operating.clean_alarm_rate.cusum.macro} & \ResultCell{operating.matched_time_fpr_at_95tpr.cusum.macro} & \ResultCell{operating.detection_delay.cusum.macro} \\
    Time-series features & \ResultCell{operating.clean_alarm_rate.time_series_features.macro} & \ResultCell{operating.matched_time_fpr_at_95tpr.time_series_features.macro} & \ResultCell{operating.detection_delay.time_series_features.macro} \\
    DEXTER & \ResultCell{operating.clean_alarm_rate.dexter.macro} & \ResultCell{operating.matched_time_fpr_at_95tpr.dexter.macro} & \ResultCell{operating.detection_delay.dexter.macro} \\
    DEXTER+C & \ResultCell{operating.clean_alarm_rate.dexter_c.macro} & \ResultCell{operating.matched_time_fpr_at_95tpr.dexter_c.macro} & \ResultCell{operating.detection_delay.dexter_c.macro} \\
    Lyapunov divergence & \ResultCell{operating.clean_alarm_rate.lyapunov_divergence.macro} & \ResultCell{operating.matched_time_fpr_at_95tpr.lyapunov_divergence.macro} & \ResultCell{operating.detection_delay.lyapunov_divergence.macro} \\
    PEDM & \ResultCell{operating.clean_alarm_rate.pedm.macro} & \ResultCell{operating.matched_time_fpr_at_95tpr.pedm.macro} & \ResultCell{operating.detection_delay.pedm.macro} \\
    STI & \ResultCell{operating.clean_alarm_rate.sti.macro} & \ResultCell{operating.matched_time_fpr_at_95tpr.sti.macro} & \ResultCell{operating.detection_delay.sti.macro} \\
    ACTI & \ResultCell{operating.clean_alarm_rate.acti.macro} & \ResultCellBest{operating.matched_time_fpr_at_95tpr.acti.macro} & \ResultCellBest{operating.detection_delay.acti.macro} \\
    Occupancy-only & \ResultCell{operating.clean_alarm_rate.occupancy_only.macro} & \ResultCell{operating.matched_time_fpr_at_95tpr.occupancy_only.macro} & \ResultCell{operating.detection_delay.occupancy_only.macro} \\
    ACTI+O & \ResultCell{operating.clean_alarm_rate.acti_occupancy.macro} & \ResultCell{operating.matched_time_fpr_at_95tpr.acti_occupancy.macro} & \ResultCell{operating.detection_delay.acti_occupancy.macro} \\
    \bottomrule
  \end{tabularx}
\normalsize
  \caption{Macro operating points on common support, averaged across all five severity tiers: clean-timestep alarm rate at the independently calibrated threshold, matched-time FPR at 95\% TPR, and detection delay in steps. Policy means with 95\% intervals. Bold marks the lowest value in the FPR and delay columns.}
  \label{tab:operating-points}
\end{table}

Table~\ref{tab:operating-points} includes the detection delay at the calibrated threshold, averaged over all five severity tiers on common support; the smoothing frontier in Figure~\ref{fig:accuracy-latency} evaluates only the severity tier $0.6$. Among the transition-model detectors, alarms are triggered fastest and within $1.5$ steps of each other. Specifically, ACTI records \Result{point.detection_delay.acti.macro} steps (utilising $L=20$ smoothing; see Section~\ref{sec:results-components}), followed by forward reconstruction error at \Result{point.detection_delay.reconstruction_error.macro}, and PEDM at \Result{point.detection_delay.pedm.macro}. The disparity between ACTI and PEDM is not statistically significant (\Result{contrast.detection_delay.acti_minus_pedm}). For window-granularity scoring, STI requires \Result{point.detection_delay.sti.macro} steps, while DEXTER requires \Result{point.detection_delay.dexter.macro}. By contrast, DEXTER+C (\Result{point.detection_delay.dexter_c.macro}), CUSUM, and Lyapunov divergence require more than one hundred steps. Relative to PEDM, ACTI enhances discrimination and operating error (AUROC \Result{contrast.matched_time_auroc.acti_minus_pedm}; FPR \Result{contrast.matched_time_fpr_at_95tpr.acti_minus_pedm}) without incurring any measurable delay cost; furthermore, a smaller smoothing window ($L\in[5,10]$) decreases latency even further (Section~\ref{sec:results-components}).

Computational cost shows a stark divide between the two detectors (Table~\ref{tab:compute-profile}). ACTI trains in \Result{point.fit_time.acti} per environment (median) and processes data at \Result{point.score_latency.acti} per transition (\Result{point.throughput.acti}) using \Result{point.parameters.acti} parameters; PEDM takes \Result{point.fit_time.pedm} to fit and scores at \Result{point.score_latency.pedm} per transition (\Result{point.throughput.pedm}) with \Result{point.parameters.pedm} parameters. Given its superior discrimination metrics, the calibrated transition detector scores roughly \Result{ratio.score_latency.pedm_over_acti} times faster and uses \Result{ratio.model_size.pedm_over_acti} times fewer parameters—suggesting potential for implementation within a control loop. The data points for DEXTER in Table~\ref{tab:compute-profile} derive from the parallel evaluation runs: tsfresh feature extraction consumes \Result{compute.tsfresh_extract_ms_range.dexter}\,ms per transition (scaling from 8-dimensional LunarLander to 128-dimensional Breakout), and a full fit across 30 trajectories requires \Result{compute.fit_minutes_range.dexter} minutes across environments. These figures place its computational burden three to five orders of magnitude higher than the single-model neural detectors.

\begin{table}[!htbp]
\centering
\footnotesize
\setlength{\tabcolsep}{2pt}
\begin{tabular*}{\textwidth}{@{\extracolsep{\fill}}lcc@{}}
\toprule
Detector & Fit (s) & Parameters \\
\midrule
Mahalanobis distance & \Result{compute.fit_seconds.mahalanobis_distance} & \Result{compute.parameters.mahalanobis_distance} \\
KDE density & \Result{compute.fit_seconds.density_kde} & \Result{compute.parameters.density_kde} \\
Ensemble disagreement & \Result{compute.fit_seconds.ensemble_disagreement} & \Result{compute.parameters.ensemble_disagreement} \\
Forward reconstruction error & \Result{compute.fit_seconds.reconstruction_error} & \Result{compute.parameters.reconstruction_error} \\
State autoencoder & \Result{compute.fit_seconds.autoencoder} & \Result{compute.parameters.autoencoder} \\
CUSUM & \Result{compute.fit_seconds.cusum} & \Result{compute.parameters.cusum} \\
Time-series features & \Result{compute.fit_seconds.time_series_features} & \Result{compute.parameters.time_series_features} \\
DEXTER & \Result{compute.fit_seconds.dexter} & --- \\
DEXTER+C & \Result{compute.fit_seconds.dexter} & --- \\
Lyapunov divergence & \Result{compute.fit_seconds.lyapunov_divergence} & \Result{compute.parameters.lyapunov_divergence} \\
PEDM & \Result{compute.fit_seconds.pedm} & \Result{compute.parameters.pedm} \\
STI & \Result{compute.fit_seconds.sti} & \Result{compute.parameters.sti} \\
ACTI & \Result{compute.fit_seconds.acti} & \Result{compute.parameters.acti} \\
Occupancy-only & \Result{compute.fit_seconds.occupancy_only} & \Result{compute.parameters.occupancy_only} \\
ACTI+O & \Result{compute.fit_seconds.acti_occupancy} & \Result{compute.parameters.acti_occupancy} \\
\bottomrule
\end{tabular*}
\par\medskip
\begin{tabular*}{\textwidth}{@{\extracolsep{\fill}}lcc@{}}
\toprule
Detector & \shortstack{Scoring time\\($\mu$s/transition)} & Throughput (transitions/s) \\
\midrule
Mahalanobis distance & \Result{compute.score_latency_us.mahalanobis_distance} & \Result{compute.throughput_tps.mahalanobis_distance} \\
KDE density & \Result{compute.score_latency_us.density_kde} & \Result{compute.throughput_tps.density_kde} \\
Ensemble disagreement & \Result{compute.score_latency_us.ensemble_disagreement} & \Result{compute.throughput_tps.ensemble_disagreement} \\
Forward reconstruction error & \Result{compute.score_latency_us.reconstruction_error} & \Result{compute.throughput_tps.reconstruction_error} \\
State autoencoder & \Result{compute.score_latency_us.autoencoder} & \Result{compute.throughput_tps.autoencoder} \\
CUSUM & \Result{compute.score_latency_us.cusum} & \Result{compute.throughput_tps.cusum} \\
Time-series features & \Result{compute.score_latency_us.time_series_features} & \Result{compute.throughput_tps.time_series_features} \\
DEXTER & \Result{compute.score_latency_us.dexter} & \Result{compute.throughput_tps.dexter} \\
DEXTER+C & \Result{compute.score_latency_us.dexter} & \Result{compute.throughput_tps.dexter} \\
Lyapunov divergence & \Result{compute.score_latency_us.lyapunov_divergence} & \Result{compute.throughput_tps.lyapunov_divergence} \\
PEDM & \Result{compute.score_latency_us.pedm} & \Result{compute.throughput_tps.pedm} \\
STI & \Result{compute.score_latency_us.sti} & \Result{compute.throughput_tps.sti} \\
ACTI & \Result{compute.score_latency_us.acti} & \Result{compute.throughput_tps.acti} \\
Occupancy-only & \Result{compute.score_latency_us.occupancy_only} & \Result{compute.throughput_tps.occupancy_only} \\
ACTI+O & \Result{compute.score_latency_us.acti_occupancy} & \Result{compute.throughput_tps.acti_occupancy} \\
\bottomrule
\end{tabular*}
\normalsize
\caption{Compute profile on saved trajectories: one fit per environment--detector cell; scoring time is complete-trajectory elapsed time divided by its transition count, using five warm-ups and 30 measured repetitions. Entries are the median [minimum, maximum] across the six environments. Peak memory is not reported because measuring it would have required a second fit per cell.}
\label{tab:compute-profile}
\end{table}

\subsection{Null diagnostics}
\label{sec:results-nulls}

The evaluation relies on two distinct categories of diagnostic checks, both of which were executed against the full fifteen-detector suite.

\textbf{Temporal-drift null.} Applying counterfactual post-onset labels to completely clean trajectories (containing no anomalies), as in the null baseline of \citet{mittag2026}, allows us to quantify the spurious detection performance that within-episode evaluation would otherwise yield. Eight of the fifteen detectors significantly distinguish these labels, including PEDM, CUSUM, the time-series and marginal-state detectors, as well as ACTI; STI, ensemble disagreement, forward reconstruction error, occupancy, ACTI+O, and the DEXTER variants remain unaffected. This specific artefact is eliminated by design through the use of matched-time negatives, which is the rationale behind their inclusion in all primary metrics.  

\textbf{Signal and channel checks.} A paired block-swap test on real matched data verifies that every detector's matched-time separation is legitimate. For each anomalous episode and its matched clean counterpart, we jointly permute the clean and anomalous labels within every non-overlapping 25-step block, while keeping the scores fixed. A null distribution for the policy-averaged macro matched-time AUROC is generated from 1,000 such draws using a single permutation stream shared across all detectors. We compare the observed value of each detector against this distribution and calculate the permutation p-value as $p = (1+k)/1001$, where $k$ represents the number of null draws that are equal to or exceed the observed statistic. After applying Holm correction, the separation is significant for all fifteen detectors; the protocol effectively filters out drift without discarding genuine signal. Reward-channel isolation operates by construction: since no detector processes reward data, pure reward drift cannot alter scores, whereas a reward-enabled CUSUM control detects this drift (\Result{null.reward_cusum_auroc_range} across severities); this invariance pertains strictly to input channels rather than an evaluation blindness. Injections with zero severity reproduce clean trajectories exactly, while action shuffling leaves the state-only detector's scores unchanged, thereby validating both label integrity and channel handling. Table~\ref{tab:nulls} outlines these five checks. None of the reported thresholds were optimised based on these results.

\begin{table}[!htbp]
\centering
\small
\begin{tabularx}{\textwidth}{@{}>{\raggedright\arraybackslash}p{3.1cm} Y >{\raggedright\arraybackslash}p{4.4cm}@{}}
\toprule
Check & What it tests & Outcome \\
\midrule
Temporal-drift null & Counterfactual post-onset labels on fully clean trajectories: does within-episode score drift alone look like detection? & \Result{null.temporal_drift_fraction} detectors separate them (incl.\ ACTI and PEDM) \\
Matched-signal check & Paired block-swap on real matched data: does each detector carry genuine matched-time signal? & \Result{null.matched_signal_fraction} significant (Holm) \\
Reward-channel isolation & Reward-only drift with all other channels fixed: do transition-detector scores move? & Scores byte-identical; reward-consuming control detects (AUROC \Result{null.reward_cusum_auroc_range}) \\
Zero-severity identity & Do zero-severity injections reproduce clean trajectories exactly? & Exact (max.\ abs.\ difference $0$) \\
Action-shuffle invariance & Do state-only detector scores change when actions are shuffled? & Byte-identical \\
\bottomrule
\end{tabularx}
\normalsize
\caption{Diagnostic checks on the evaluation protocol. The first check measures an artefact of \emph{uncontrolled} evaluation (which matched-time negatives remove); the remaining four validate that the controlled protocol preserves real signal and respects input channels.}
\label{tab:nulls}
\end{table}

\section{Discussion}
\label{sec:discussion}

\subsection{Interpreting the ACTI-Forward comparison}
\label{sec:discussion-null}

The contrast was originally designed to counteract prediction difficulty shared across both temporal directions, thereby isolating anomalous asymmetry. Two factors may explain the lack of additional benefit. Firstly, per-timestep calibration removes each statistic's baseline profile, which may already account for part of the shared prediction difficulty that the contrast was intended to neutralise. Consequently, the contrast and the calibration may act as overlapping solutions to the same problem. Secondly, given the conditional-descriptor convention ($\bar a_t=a_t$), both models are trained on identical transitions with the same action descriptor; their residuals may therefore be correlated, which could limit the additional information provided by the contrast.

\subsection{The effect of action conditioning}

A state-only transition model marginalises over the policy’s action distribution. In scenarios where multiple actions are plausible within similar states yet generate distinct successors, this averaged model becomes broad or multimodal, allowing anomalous successors to remain concealed within its predicted range. Conditioning on the recorded commanded action collapses this mixture, and the empirical results align with these predictions: a modest but consistent improvement in discrimination (\Result{contrast.matched_time_auroc.acti_minus_sti}), a notable reduction in detection delay (\Result{point.contrast_delay.acti_minus_sti_signed} steps relative to STI, consistent with sharper per-step statistics reaching the threshold more rapidly), and a lower false-positive rate at 95\% TPR. This benefit remains relatively limited, consistent with the evaluated policies being predominantly deterministic given the state (with only minor exploration noise during evaluation), which restricts the additional information the action provides beyond the state itself. Implementations characterised by higher action entropy might amplify this gap. While testing this scenario is feasible, it requires retraining policies and requalifying severity tiers; consequently, it falls outside the scope of the present study.

\subsection{Operating characteristics beyond AUROC}

Three detectors illustrate practical trade-offs. PEDM alarms after \Result{point.detection_delay.pedm.macro} steps but requires \Result{point.score_latency.pedm} per transition and underperforms ACTI across all discrimination metrics. Forward reconstruction error operates at a comparable detection delay (\Result{point.detection_delay.reconstruction_error.approx} steps), incurs minimal computational overhead (\Result{point.score_latency.reconstruction_error}), and delivers intermediate discrimination performance. ACTI attains a macro matched-time AUROC of \Result{main.matched_time_auroc.acti.macro}, a detection delay of \Result{point.detection_delay.acti.macro} steps, and a scoring time of \Result{point.score_latency.acti} per transition. The ACTI-Forward baseline exhibited a significantly higher AUPRC alongside reduced detection delay, without presenting any statistically meaningful divergence in either AUROC or FPR. Utilising a narrower ACTI smoothing window ($L=5$) reduces detection delay while preserving the majority of the accuracy. The calibrated detector demonstrates its greatest advantage at lower anomaly tiers (\Result{severity.tier1.matched_time_auroc.acti} versus \Result{severity.tier1.matched_time_auroc.pedm} macro AUROC at the lowest tier). DEXTER trails ACTI across every discrimination metric, scores roughly four orders of magnitude slower, and triggers alarms later (\Result{point.detection_delay.dexter.macro} steps). For deployment within a control loop, the potential value of this transition-likelihood approach lies in its combination of leading discrimination performance, microsecond-scale scoring, and a tunable accuracy–delay trade-off, independent of the physical interpretation that originally motivated it.

\subsection{Statistical irreversibility is not entropy production}
\label{sec:discussion-thermo}

While research into fluctuation theorems motivated the use of forward and reverse comparisons, a thermodynamic interpretation of ACTI requires a distinct derivation. Our experimental setup includes policies featuring both discrete and continuous actions, partially observed simulator dynamics, and faults where commanded and executed actions diverge. We have not outlined the physical time-reversal transformation or conjugate path measure required to link these controlled trajectories to entropy production. Furthermore, ACTI relies on independently learned one-step conditional densities, empirical centring, absolute deviations, and temporal smoothing techniques. Partial observation alone does not preclude an entropy-production bound, nor does the lack of an externally reversed protocol constitute a general obstruction for stationary processes \citep{seara2021}. We refrain from providing a physical interpretation due to the absence of a derivation for this specific statistic and these systems. This constraint remains applicable regardless of the ACTI-Forward comparison. Consequently, we describe the evaluated detectors as calibrated, action-conditioned transition-likelihood monitors.

\subsection{Implications for future OOD evaluation in RL}

For future RL-OOD research, we recommend comparing anomalous and clean scores at matched timesteps, and selecting anomaly severities independent of detector performance. In our matched-time evaluation with multiple policies, the increase in AUROC from action conditioning was an order of magnitude smaller than that observed in the single-policy, within-episode pilot. Furthermore, the drift null metric demonstrates that eight out of fifteen detectors can distinguish counterfactual labels on entirely clean data, illustrating how conventional within-episode evaluation can inflate detection performance estimates. Detector-blind severity qualification avoided potential bias from selecting severity thresholds that favour particular methods, although this necessitated reducing the comparable anomaly set and excluding scenarios where effects could not be adequately controlled (refer to Table~\ref{tab:qualification-map}). Without incorporating the capacity-matched ACTI-Forward control, we would have incorrectly attributed our superior ranking to the forward--inverse likelihood contrast.

\section{Limitations}
\label{sec:limitations}
Firstly, primary inference relies on a detector-blind common support framework, admitting four mechanisms across all six environments. Of the 54 screened environment-anomaly combinations, 24 constitute the common support set. A further fifteen qualified within specific environments only and are documented as environment-specific evidence, while the final fifteen failed qualification for reasons detailed in Table~\ref{tab:qualification-map}. Consequently, the study yields no conclusions regarding the excluded cells; specifically, observation delay and action-conditioned dynamics remain untested in four of the six environments. The minimum-common-support threshold was subsequently relaxed from six to four mechanisms following the screening outcome, yet before any detector evaluation commenced.

Secondly, severity tiers function ordinally as per-anomaly effect levels, having been qualified on a single pilot policy. They do not equate absolute effect sizes across different anomalies or environments, nor do they correspond to realised effects under the evaluated policies.

Thirdly, all six testbeds are simulated environments, meaning the online dynamics faults serve as proxy interventions with fidelity that varies considerably across settings (ranging from full state-level interventions in MuJoCo tasks to observation-level feedback in Breakout). The setup does not capture real-world sensor noise, hardware degradation, or compound faults, and anomaly onsets are strictly abrupt and predetermined.

Fourthly, learned reverse dynamics may introduce greater ambiguity compared to forward dynamics, and the Gaussian residual model might understate such multimodality. While a more expressive reverse density could theoretically restore the contrast's utility, it would simultaneously elevate the performance threshold for forward-only models.

Finally, the Breakout findings rely upon a single fixed random projection; the score-transformation ablation leaves the state representation unchanged.

\section{Broader Impact and Safety Considerations}
\label{sec:broader-impact}

Out-of-distribution alarms can assist reinforcement learning systems in deferring, halting, or switching to a fallback controller when deployment conditions shift, yet they cannot inherently guarantee safety. False negatives may permit hazardous operations, while false positives trigger unnecessary shutdowns or train operators to disregard alerts. For this reason, we present explicit operating points, latency figures, and computational requirements alongside discrimination metrics, while cautioning that performance at ordinal severity tiers does not equate to direct fault-magnitude guarantees. Although open-sourcing our code, checkpoints, and comprehensive results enhances reproducibility, it simultaneously allows subsequent detectors to overfit the publicly available anomaly suite; therefore, evaluating against held-out anomaly families remains the most appropriate defence.

\section{Conclusion}
\label{sec:conclusion}

This study investigated whether action-conditioned transition irreversibility—a forward-versus-reverse transition likelihood contrast—serves as a reliable out-of-distribution signal for reinforcement learning trajectories. To mitigate the failure modes that typically inflate RL-OOD results, our evaluation utilised matched-time negatives, detector-blind severity tiers, ten policy replicates per environment, and pre-specified capacity-matched controls. Across a comprehensive comparison spanning fifteen detectors and six environments, the resulting model achieved the highest macro matched-time AUROC (\Result{main.matched_time_auroc.acti.macro}). It substantially surpassed PEDM in both discrimination metrics and operating-point error rates, with approximately two orders of magnitude lower scoring time. Its performance advantage remains most pronounced when detecting subtle anomalies. 

These experiments failed to demonstrate any additional detection advantage derived from the forward--inverse likelihood contrast under the evaluated conditions. ACTI-Forward attained a significantly higher AUPRC alongside a reduced detection delay, whereas AUROC and FPR comparisons revealed no statistically significant divergence. Ablation studies support contributions from action conditioning, per-timestep clean calibration, and causal smoothing applied to the transition likelihood. We therefore present these detectors as a practical advance while treating the mechanism-level null result with equal significance.  Our evaluation protocol is necessarily rigorous; had we relied on the within-episode pilot alone, we would have reported an AUROC gain from action conditioning that was an order of magnitude larger than the gain we observed from the main evaluation. This methodological framework underpins our confidence in both the positive findings and the negative results. Our results indicate that ACTI-Forward is the method of choice among the detectors evaluated.

\section*{Acknowledgements}
This research was supported by research funding from Deakin University.

\bibliography{references}

\clearpage
\appendix
\section{Matched-time AUPRC results}
\label{app:auprc-results}

\begin{table}[!htbp]
  \centering
  \footnotesize
  \setlength{\tabcolsep}{2pt}
  \begin{tabular*}{\textwidth}{@{\extracolsep{\fill}}lcccc@{}}
    \toprule
    Detector & LunarLander & Acrobot & HalfCheetah & Ant \\
    \midrule
    Mahalanobis distance & \ResultCell{main.matched_time_auprc.mahalanobis_distance.ll} & \ResultCell{main.matched_time_auprc.mahalanobis_distance.ac} & \ResultCell{main.matched_time_auprc.mahalanobis_distance.hc} & \ResultCell{main.matched_time_auprc.mahalanobis_distance.an} \\
    KDE density & \ResultCell{main.matched_time_auprc.density_kde.ll} & \ResultCell{main.matched_time_auprc.density_kde.ac} & \ResultCell{main.matched_time_auprc.density_kde.hc} & \ResultCell{main.matched_time_auprc.density_kde.an} \\
    Ensemble disagreement & \ResultCell{main.matched_time_auprc.ensemble_disagreement.ll} & \ResultCell{main.matched_time_auprc.ensemble_disagreement.ac} & \ResultCell{main.matched_time_auprc.ensemble_disagreement.hc} & \ResultCell{main.matched_time_auprc.ensemble_disagreement.an} \\
    Forward reconstruction error & \ResultCell{main.matched_time_auprc.reconstruction_error.ll} & \ResultCell{main.matched_time_auprc.reconstruction_error.ac} & \ResultCell{main.matched_time_auprc.reconstruction_error.hc} & \ResultCell{main.matched_time_auprc.reconstruction_error.an} \\
    State autoencoder & \ResultCell{main.matched_time_auprc.autoencoder.ll} & \ResultCell{main.matched_time_auprc.autoencoder.ac} & \ResultCell{main.matched_time_auprc.autoencoder.hc} & \ResultCell{main.matched_time_auprc.autoencoder.an} \\
    CUSUM & \ResultCell{main.matched_time_auprc.cusum.ll} & \ResultCell{main.matched_time_auprc.cusum.ac} & \ResultCell{main.matched_time_auprc.cusum.hc} & \ResultCell{main.matched_time_auprc.cusum.an} \\
    Time-series features & \ResultCell{main.matched_time_auprc.time_series_features.ll} & \ResultCell{main.matched_time_auprc.time_series_features.ac} & \ResultCell{main.matched_time_auprc.time_series_features.hc} & \ResultCell{main.matched_time_auprc.time_series_features.an} \\
    DEXTER & \ResultCell{main.matched_time_auprc.dexter.ll} & \ResultCell{main.matched_time_auprc.dexter.ac} & \ResultCell{main.matched_time_auprc.dexter.hc} & \ResultCell{main.matched_time_auprc.dexter.an} \\
    DEXTER+C & \ResultCell{main.matched_time_auprc.dexter_c.ll} & \ResultCell{main.matched_time_auprc.dexter_c.ac} & \ResultCell{main.matched_time_auprc.dexter_c.hc} & \ResultCell{main.matched_time_auprc.dexter_c.an} \\
    Lyapunov divergence & \ResultCell{main.matched_time_auprc.lyapunov_divergence.ll} & \ResultCell{main.matched_time_auprc.lyapunov_divergence.ac} & \ResultCell{main.matched_time_auprc.lyapunov_divergence.hc} & \ResultCell{main.matched_time_auprc.lyapunov_divergence.an} \\
    PEDM & \ResultCell{main.matched_time_auprc.pedm.ll} & \ResultCell{main.matched_time_auprc.pedm.ac} & \ResultCell{main.matched_time_auprc.pedm.hc} & \ResultCell{main.matched_time_auprc.pedm.an} \\
    STI & \ResultCellBest{main.matched_time_auprc.sti.ll} & \ResultCell{main.matched_time_auprc.sti.ac} & \ResultCell{main.matched_time_auprc.sti.hc} & \ResultCell{main.matched_time_auprc.sti.an} \\
    ACTI & \ResultCell{main.matched_time_auprc.acti.ll} & \ResultCell{main.matched_time_auprc.acti.ac} & \ResultCellBest{main.matched_time_auprc.acti.hc} & \ResultCellBest{main.matched_time_auprc.acti.an} \\
    Occupancy-only & \ResultCell{main.matched_time_auprc.occupancy_only.ll} & \ResultCell{main.matched_time_auprc.occupancy_only.ac} & \ResultCell{main.matched_time_auprc.occupancy_only.hc} & \ResultCell{main.matched_time_auprc.occupancy_only.an} \\
    ACTI+O & \ResultCell{main.matched_time_auprc.acti_occupancy.ll} & \ResultCellBest{main.matched_time_auprc.acti_occupancy.ac} & \ResultCell{main.matched_time_auprc.acti_occupancy.hc} & \ResultCell{main.matched_time_auprc.acti_occupancy.an} \\
    \bottomrule
  \end{tabular*}
  \par\medskip
  \begin{tabular*}{\textwidth}{@{\extracolsep{\fill}}lccc@{}}
    \toprule
    Detector & Humanoid & Breakout & Macro mean \\
    \midrule
    Mahalanobis distance & \ResultCell{main.matched_time_auprc.mahalanobis_distance.hu} & \ResultCell{main.matched_time_auprc.mahalanobis_distance.br} & \ResultCell{main.matched_time_auprc.mahalanobis_distance.macro} \\
    KDE density & \ResultCell{main.matched_time_auprc.density_kde.hu} & \ResultCell{main.matched_time_auprc.density_kde.br} & \ResultCell{main.matched_time_auprc.density_kde.macro} \\
    Ensemble disagreement & \ResultCell{main.matched_time_auprc.ensemble_disagreement.hu} & \ResultCell{main.matched_time_auprc.ensemble_disagreement.br} & \ResultCell{main.matched_time_auprc.ensemble_disagreement.macro} \\
    Forward reconstruction error & \ResultCell{main.matched_time_auprc.reconstruction_error.hu} & \ResultCell{main.matched_time_auprc.reconstruction_error.br} & \ResultCell{main.matched_time_auprc.reconstruction_error.macro} \\
    State autoencoder & \ResultCell{main.matched_time_auprc.autoencoder.hu} & \ResultCell{main.matched_time_auprc.autoencoder.br} & \ResultCell{main.matched_time_auprc.autoencoder.macro} \\
    CUSUM & \ResultCell{main.matched_time_auprc.cusum.hu} & \ResultCell{main.matched_time_auprc.cusum.br} & \ResultCell{main.matched_time_auprc.cusum.macro} \\
    Time-series features & \ResultCell{main.matched_time_auprc.time_series_features.hu} & \ResultCell{main.matched_time_auprc.time_series_features.br} & \ResultCell{main.matched_time_auprc.time_series_features.macro} \\
    DEXTER & \ResultCell{main.matched_time_auprc.dexter.hu} & \ResultCell{main.matched_time_auprc.dexter.br} & \ResultCell{main.matched_time_auprc.dexter.macro} \\
    DEXTER+C & \ResultCell{main.matched_time_auprc.dexter_c.hu} & \ResultCell{main.matched_time_auprc.dexter_c.br} & \ResultCell{main.matched_time_auprc.dexter_c.macro} \\
    Lyapunov divergence & \ResultCell{main.matched_time_auprc.lyapunov_divergence.hu} & \ResultCell{main.matched_time_auprc.lyapunov_divergence.br} & \ResultCell{main.matched_time_auprc.lyapunov_divergence.macro} \\
    PEDM & \ResultCell{main.matched_time_auprc.pedm.hu} & \ResultCell{main.matched_time_auprc.pedm.br} & \ResultCell{main.matched_time_auprc.pedm.macro} \\
    STI & \ResultCell{main.matched_time_auprc.sti.hu} & \ResultCell{main.matched_time_auprc.sti.br} & \ResultCell{main.matched_time_auprc.sti.macro} \\
    ACTI & \ResultCellBest{main.matched_time_auprc.acti.hu} & \ResultCellBest{main.matched_time_auprc.acti.br} & \ResultCellBest{main.matched_time_auprc.acti.macro} \\
    Occupancy-only & \ResultCell{main.matched_time_auprc.occupancy_only.hu} & \ResultCell{main.matched_time_auprc.occupancy_only.br} & \ResultCell{main.matched_time_auprc.occupancy_only.macro} \\
    ACTI+O & \ResultCell{main.matched_time_auprc.acti_occupancy.hu} & \ResultCell{main.matched_time_auprc.acti_occupancy.br} & \ResultCell{main.matched_time_auprc.acti_occupancy.macro} \\
    \bottomrule
  \end{tabular*}
\normalsize
  \caption{Policy-level matched-time AUPRC on the frozen common support $A^*$ (four mechanisms, five tiers, tier-averaged), mean and 95\% policy-bootstrap interval. Macro weights environments equally. Bold marks the highest value in each column.}
  \label{tab:auprc-results}
\end{table}

\clearpage
\section{Expanded-support AUROC results}
\label{app:expanded-auroc-results}

\begingroup
\footnotesize
\setlength{\tabcolsep}{4pt}
\setlength{\LTleft}{\fill}
\setlength{\LTright}{\fill}
\setlength{\LTcapwidth}{\textwidth}
\begin{longtable}{@{}p{0.29\textwidth}*{3}{>{\centering\arraybackslash}p{0.21\textwidth}}@{}}
\endfirsthead
\multicolumn{4}{l}{Table~\thetable{} (continued)} \\
\addlinespace[0.5em]
\endhead
\midrule
\multicolumn{4}{r}{Continued on next page} \\
\endfoot
\bottomrule
\caption{Matched-time AUROC for the fifteen environment-specific qualified cells outside common support, averaged across all five severity tiers. Entries show the mean and 95\% policy-bootstrap interval across ten policies per environment. Bold marks the highest point estimate among the fifteen suite detectors in each column.}\label{tab:expanded-auroc-results} \\
\endlastfoot
\toprule
Detector & \shortstack{LunarLander\\Correlation break} & \shortstack{LunarLander\\Dynamics scaling} & \shortstack{LunarLander\\Gradual drift} \\*
\midrule
Mahalanobis distance & \ResultCell{expanded_table.matched_time_auroc.mahalanobis_distance.lunar_lander.correlation_break} & \ResultCell{expanded_table.matched_time_auroc.mahalanobis_distance.lunar_lander.dynamics_scaling} & \ResultCell{expanded_table.matched_time_auroc.mahalanobis_distance.lunar_lander.gradual_drift} \\*
KDE density & \ResultCell{expanded_table.matched_time_auroc.density_kde.lunar_lander.correlation_break} & \ResultCell{expanded_table.matched_time_auroc.density_kde.lunar_lander.dynamics_scaling} & \ResultCell{expanded_table.matched_time_auroc.density_kde.lunar_lander.gradual_drift} \\*
Ensemble disagreement & \ResultCell{expanded_table.matched_time_auroc.ensemble_disagreement.lunar_lander.correlation_break} & \ResultCell{expanded_table.matched_time_auroc.ensemble_disagreement.lunar_lander.dynamics_scaling} & \ResultCell{expanded_table.matched_time_auroc.ensemble_disagreement.lunar_lander.gradual_drift} \\*
Forward reconstruction error & \ResultCell{expanded_table.matched_time_auroc.reconstruction_error.lunar_lander.correlation_break} & \ResultCell{expanded_table.matched_time_auroc.reconstruction_error.lunar_lander.dynamics_scaling} & \ResultCell{expanded_table.matched_time_auroc.reconstruction_error.lunar_lander.gradual_drift} \\*
State autoencoder & \ResultCell{expanded_table.matched_time_auroc.autoencoder.lunar_lander.correlation_break} & \ResultCell{expanded_table.matched_time_auroc.autoencoder.lunar_lander.dynamics_scaling} & \ResultCell{expanded_table.matched_time_auroc.autoencoder.lunar_lander.gradual_drift} \\*
CUSUM & \ResultCell{expanded_table.matched_time_auroc.cusum.lunar_lander.correlation_break} & \ResultCell{expanded_table.matched_time_auroc.cusum.lunar_lander.dynamics_scaling} & \ResultCell{expanded_table.matched_time_auroc.cusum.lunar_lander.gradual_drift} \\*
Time-series features & \ResultCell{expanded_table.matched_time_auroc.time_series_features.lunar_lander.correlation_break} & \ResultCell{expanded_table.matched_time_auroc.time_series_features.lunar_lander.dynamics_scaling} & \ResultCell{expanded_table.matched_time_auroc.time_series_features.lunar_lander.gradual_drift} \\*
DEXTER & \ResultCell{expanded_table.matched_time_auroc.dexter.lunar_lander.correlation_break} & \ResultCell{expanded_table.matched_time_auroc.dexter.lunar_lander.dynamics_scaling} & \ResultCell{expanded_table.matched_time_auroc.dexter.lunar_lander.gradual_drift} \\*
DEXTER+C & \ResultCell{expanded_table.matched_time_auroc.dexter_c.lunar_lander.correlation_break} & \ResultCell{expanded_table.matched_time_auroc.dexter_c.lunar_lander.dynamics_scaling} & \ResultCell{expanded_table.matched_time_auroc.dexter_c.lunar_lander.gradual_drift} \\*
Lyapunov divergence & \ResultCell{expanded_table.matched_time_auroc.lyapunov_divergence.lunar_lander.correlation_break} & \ResultCell{expanded_table.matched_time_auroc.lyapunov_divergence.lunar_lander.dynamics_scaling} & \ResultCell{expanded_table.matched_time_auroc.lyapunov_divergence.lunar_lander.gradual_drift} \\*
PEDM & \ResultCell{expanded_table.matched_time_auroc.pedm.lunar_lander.correlation_break} & \ResultCell{expanded_table.matched_time_auroc.pedm.lunar_lander.dynamics_scaling} & \ResultCellBest{expanded_table.matched_time_auroc.pedm.lunar_lander.gradual_drift} \\*
STI & \ResultCell{expanded_table.matched_time_auroc.sti.lunar_lander.correlation_break} & \ResultCell{expanded_table.matched_time_auroc.sti.lunar_lander.dynamics_scaling} & \ResultCell{expanded_table.matched_time_auroc.sti.lunar_lander.gradual_drift} \\*
ACTI & \ResultCellBest{expanded_table.matched_time_auroc.acti.lunar_lander.correlation_break} & \ResultCellBest{expanded_table.matched_time_auroc.acti.lunar_lander.dynamics_scaling} & \ResultCell{expanded_table.matched_time_auroc.acti.lunar_lander.gradual_drift} \\*
Occupancy-only & \ResultCell{expanded_table.matched_time_auroc.occupancy_only.lunar_lander.correlation_break} & \ResultCell{expanded_table.matched_time_auroc.occupancy_only.lunar_lander.dynamics_scaling} & \ResultCell{expanded_table.matched_time_auroc.occupancy_only.lunar_lander.gradual_drift} \\*
ACTI+O & \ResultCell{expanded_table.matched_time_auroc.acti_occupancy.lunar_lander.correlation_break} & \ResultCell{expanded_table.matched_time_auroc.acti_occupancy.lunar_lander.dynamics_scaling} & \ResultCell{expanded_table.matched_time_auroc.acti_occupancy.lunar_lander.gradual_drift} \\
\addlinespace[1em]
\toprule
Detector & \shortstack{LunarLander\\Observation delay} & \shortstack{Acrobot\\Dynamics scaling} & \shortstack{Acrobot\\Gradual drift} \\*
\midrule
Mahalanobis distance & \ResultCell{expanded_table.matched_time_auroc.mahalanobis_distance.lunar_lander.observation_delay} & \ResultCell{expanded_table.matched_time_auroc.mahalanobis_distance.acrobot.dynamics_scaling} & \ResultCell{expanded_table.matched_time_auroc.mahalanobis_distance.acrobot.gradual_drift} \\*
KDE density & \ResultCell{expanded_table.matched_time_auroc.density_kde.lunar_lander.observation_delay} & \ResultCell{expanded_table.matched_time_auroc.density_kde.acrobot.dynamics_scaling} & \ResultCell{expanded_table.matched_time_auroc.density_kde.acrobot.gradual_drift} \\*
Ensemble disagreement & \ResultCell{expanded_table.matched_time_auroc.ensemble_disagreement.lunar_lander.observation_delay} & \ResultCell{expanded_table.matched_time_auroc.ensemble_disagreement.acrobot.dynamics_scaling} & \ResultCell{expanded_table.matched_time_auroc.ensemble_disagreement.acrobot.gradual_drift} \\*
Forward reconstruction error & \ResultCell{expanded_table.matched_time_auroc.reconstruction_error.lunar_lander.observation_delay} & \ResultCell{expanded_table.matched_time_auroc.reconstruction_error.acrobot.dynamics_scaling} & \ResultCell{expanded_table.matched_time_auroc.reconstruction_error.acrobot.gradual_drift} \\*
State autoencoder & \ResultCell{expanded_table.matched_time_auroc.autoencoder.lunar_lander.observation_delay} & \ResultCell{expanded_table.matched_time_auroc.autoencoder.acrobot.dynamics_scaling} & \ResultCell{expanded_table.matched_time_auroc.autoencoder.acrobot.gradual_drift} \\*
CUSUM & \ResultCell{expanded_table.matched_time_auroc.cusum.lunar_lander.observation_delay} & \ResultCell{expanded_table.matched_time_auroc.cusum.acrobot.dynamics_scaling} & \ResultCell{expanded_table.matched_time_auroc.cusum.acrobot.gradual_drift} \\*
Time-series features & \ResultCell{expanded_table.matched_time_auroc.time_series_features.lunar_lander.observation_delay} & \ResultCell{expanded_table.matched_time_auroc.time_series_features.acrobot.dynamics_scaling} & \ResultCell{expanded_table.matched_time_auroc.time_series_features.acrobot.gradual_drift} \\*
DEXTER & \ResultCell{expanded_table.matched_time_auroc.dexter.lunar_lander.observation_delay} & \ResultCell{expanded_table.matched_time_auroc.dexter.acrobot.dynamics_scaling} & \ResultCell{expanded_table.matched_time_auroc.dexter.acrobot.gradual_drift} \\*
DEXTER+C & \ResultCell{expanded_table.matched_time_auroc.dexter_c.lunar_lander.observation_delay} & \ResultCell{expanded_table.matched_time_auroc.dexter_c.acrobot.dynamics_scaling} & \ResultCell{expanded_table.matched_time_auroc.dexter_c.acrobot.gradual_drift} \\*
Lyapunov divergence & \ResultCell{expanded_table.matched_time_auroc.lyapunov_divergence.lunar_lander.observation_delay} & \ResultCell{expanded_table.matched_time_auroc.lyapunov_divergence.acrobot.dynamics_scaling} & \ResultCell{expanded_table.matched_time_auroc.lyapunov_divergence.acrobot.gradual_drift} \\*
PEDM & \ResultCellBest{expanded_table.matched_time_auroc.pedm.lunar_lander.observation_delay} & \ResultCell{expanded_table.matched_time_auroc.pedm.acrobot.dynamics_scaling} & \ResultCell{expanded_table.matched_time_auroc.pedm.acrobot.gradual_drift} \\*
STI & \ResultCell{expanded_table.matched_time_auroc.sti.lunar_lander.observation_delay} & \ResultCell{expanded_table.matched_time_auroc.sti.acrobot.dynamics_scaling} & \ResultCell{expanded_table.matched_time_auroc.sti.acrobot.gradual_drift} \\*
ACTI & \ResultCell{expanded_table.matched_time_auroc.acti.lunar_lander.observation_delay} & \ResultCellBest{expanded_table.matched_time_auroc.acti.acrobot.dynamics_scaling} & \ResultCellBest{expanded_table.matched_time_auroc.acti.acrobot.gradual_drift} \\*
Occupancy-only & \ResultCell{expanded_table.matched_time_auroc.occupancy_only.lunar_lander.observation_delay} & \ResultCell{expanded_table.matched_time_auroc.occupancy_only.acrobot.dynamics_scaling} & \ResultCell{expanded_table.matched_time_auroc.occupancy_only.acrobot.gradual_drift} \\*
ACTI+O & \ResultCell{expanded_table.matched_time_auroc.acti_occupancy.lunar_lander.observation_delay} & \ResultCell{expanded_table.matched_time_auroc.acti_occupancy.acrobot.dynamics_scaling} & \ResultCell{expanded_table.matched_time_auroc.acti_occupancy.acrobot.gradual_drift} \\
\pagebreak
\toprule
Detector & \shortstack{Acrobot\\Action-conditioned\\dynamics} & \shortstack{HalfCheetah\\Correlation break} & \shortstack{Ant\\Correlation break} \\*
\midrule
Mahalanobis distance & \ResultCell{expanded_table.matched_time_auroc.mahalanobis_distance.acrobot.action_conditioned_dynamics} & \ResultCell{expanded_table.matched_time_auroc.mahalanobis_distance.half_cheetah.correlation_break} & \ResultCell{expanded_table.matched_time_auroc.mahalanobis_distance.ant.correlation_break} \\*
KDE density & \ResultCell{expanded_table.matched_time_auroc.density_kde.acrobot.action_conditioned_dynamics} & \ResultCell{expanded_table.matched_time_auroc.density_kde.half_cheetah.correlation_break} & \ResultCell{expanded_table.matched_time_auroc.density_kde.ant.correlation_break} \\*
Ensemble disagreement & \ResultCell{expanded_table.matched_time_auroc.ensemble_disagreement.acrobot.action_conditioned_dynamics} & \ResultCell{expanded_table.matched_time_auroc.ensemble_disagreement.half_cheetah.correlation_break} & \ResultCell{expanded_table.matched_time_auroc.ensemble_disagreement.ant.correlation_break} \\*
Forward reconstruction error & \ResultCell{expanded_table.matched_time_auroc.reconstruction_error.acrobot.action_conditioned_dynamics} & \ResultCell{expanded_table.matched_time_auroc.reconstruction_error.half_cheetah.correlation_break} & \ResultCell{expanded_table.matched_time_auroc.reconstruction_error.ant.correlation_break} \\*
State autoencoder & \ResultCell{expanded_table.matched_time_auroc.autoencoder.acrobot.action_conditioned_dynamics} & \ResultCell{expanded_table.matched_time_auroc.autoencoder.half_cheetah.correlation_break} & \ResultCell{expanded_table.matched_time_auroc.autoencoder.ant.correlation_break} \\*
CUSUM & \ResultCell{expanded_table.matched_time_auroc.cusum.acrobot.action_conditioned_dynamics} & \ResultCell{expanded_table.matched_time_auroc.cusum.half_cheetah.correlation_break} & \ResultCell{expanded_table.matched_time_auroc.cusum.ant.correlation_break} \\*
Time-series features & \ResultCell{expanded_table.matched_time_auroc.time_series_features.acrobot.action_conditioned_dynamics} & \ResultCell{expanded_table.matched_time_auroc.time_series_features.half_cheetah.correlation_break} & \ResultCell{expanded_table.matched_time_auroc.time_series_features.ant.correlation_break} \\*
DEXTER & \ResultCell{expanded_table.matched_time_auroc.dexter.acrobot.action_conditioned_dynamics} & \ResultCell{expanded_table.matched_time_auroc.dexter.half_cheetah.correlation_break} & \ResultCell{expanded_table.matched_time_auroc.dexter.ant.correlation_break} \\*
DEXTER+C & \ResultCell{expanded_table.matched_time_auroc.dexter_c.acrobot.action_conditioned_dynamics} & \ResultCell{expanded_table.matched_time_auroc.dexter_c.half_cheetah.correlation_break} & \ResultCell{expanded_table.matched_time_auroc.dexter_c.ant.correlation_break} \\*
Lyapunov divergence & \ResultCell{expanded_table.matched_time_auroc.lyapunov_divergence.acrobot.action_conditioned_dynamics} & \ResultCell{expanded_table.matched_time_auroc.lyapunov_divergence.half_cheetah.correlation_break} & \ResultCell{expanded_table.matched_time_auroc.lyapunov_divergence.ant.correlation_break} \\*
PEDM & \ResultCell{expanded_table.matched_time_auroc.pedm.acrobot.action_conditioned_dynamics} & \ResultCell{expanded_table.matched_time_auroc.pedm.half_cheetah.correlation_break} & \ResultCell{expanded_table.matched_time_auroc.pedm.ant.correlation_break} \\*
STI & \ResultCell{expanded_table.matched_time_auroc.sti.acrobot.action_conditioned_dynamics} & \ResultCellBest{expanded_table.matched_time_auroc.sti.half_cheetah.correlation_break} & \ResultCell{expanded_table.matched_time_auroc.sti.ant.correlation_break} \\*
ACTI & \ResultCell{expanded_table.matched_time_auroc.acti.acrobot.action_conditioned_dynamics} & \ResultCell{expanded_table.matched_time_auroc.acti.half_cheetah.correlation_break} & \ResultCellBest{expanded_table.matched_time_auroc.acti.ant.correlation_break} \\*
Occupancy-only & \ResultCell{expanded_table.matched_time_auroc.occupancy_only.acrobot.action_conditioned_dynamics} & \ResultCell{expanded_table.matched_time_auroc.occupancy_only.half_cheetah.correlation_break} & \ResultCell{expanded_table.matched_time_auroc.occupancy_only.ant.correlation_break} \\*
ACTI+O & \ResultCellBest{expanded_table.matched_time_auroc.acti_occupancy.acrobot.action_conditioned_dynamics} & \ResultCell{expanded_table.matched_time_auroc.acti_occupancy.half_cheetah.correlation_break} & \ResultCell{expanded_table.matched_time_auroc.acti_occupancy.ant.correlation_break} \\
\addlinespace[1em]
\toprule
Detector & \shortstack{Humanoid\\Correlation break} & \shortstack{Breakout\\Correlation break} & \shortstack{Breakout\\Dynamics scaling} \\*
\midrule
Mahalanobis distance & \ResultCell{expanded_table.matched_time_auroc.mahalanobis_distance.humanoid.correlation_break} & \ResultCell{expanded_table.matched_time_auroc.mahalanobis_distance.breakout.correlation_break} & \ResultCell{expanded_table.matched_time_auroc.mahalanobis_distance.breakout.dynamics_scaling} \\*
KDE density & \ResultCell{expanded_table.matched_time_auroc.density_kde.humanoid.correlation_break} & \ResultCell{expanded_table.matched_time_auroc.density_kde.breakout.correlation_break} & \ResultCell{expanded_table.matched_time_auroc.density_kde.breakout.dynamics_scaling} \\*
Ensemble disagreement & \ResultCell{expanded_table.matched_time_auroc.ensemble_disagreement.humanoid.correlation_break} & \ResultCell{expanded_table.matched_time_auroc.ensemble_disagreement.breakout.correlation_break} & \ResultCell{expanded_table.matched_time_auroc.ensemble_disagreement.breakout.dynamics_scaling} \\*
Forward reconstruction error & \ResultCell{expanded_table.matched_time_auroc.reconstruction_error.humanoid.correlation_break} & \ResultCell{expanded_table.matched_time_auroc.reconstruction_error.breakout.correlation_break} & \ResultCell{expanded_table.matched_time_auroc.reconstruction_error.breakout.dynamics_scaling} \\*
State autoencoder & \ResultCell{expanded_table.matched_time_auroc.autoencoder.humanoid.correlation_break} & \ResultCell{expanded_table.matched_time_auroc.autoencoder.breakout.correlation_break} & \ResultCell{expanded_table.matched_time_auroc.autoencoder.breakout.dynamics_scaling} \\*
CUSUM & \ResultCell{expanded_table.matched_time_auroc.cusum.humanoid.correlation_break} & \ResultCell{expanded_table.matched_time_auroc.cusum.breakout.correlation_break} & \ResultCell{expanded_table.matched_time_auroc.cusum.breakout.dynamics_scaling} \\*
Time-series features & \ResultCell{expanded_table.matched_time_auroc.time_series_features.humanoid.correlation_break} & \ResultCell{expanded_table.matched_time_auroc.time_series_features.breakout.correlation_break} & \ResultCellBest{expanded_table.matched_time_auroc.time_series_features.breakout.dynamics_scaling} \\*
DEXTER & \ResultCell{expanded_table.matched_time_auroc.dexter.humanoid.correlation_break} & \ResultCell{expanded_table.matched_time_auroc.dexter.breakout.correlation_break} & \ResultCell{expanded_table.matched_time_auroc.dexter.breakout.dynamics_scaling} \\*
DEXTER+C & \ResultCell{expanded_table.matched_time_auroc.dexter_c.humanoid.correlation_break} & \ResultCell{expanded_table.matched_time_auroc.dexter_c.breakout.correlation_break} & \ResultCell{expanded_table.matched_time_auroc.dexter_c.breakout.dynamics_scaling} \\*
Lyapunov divergence & \ResultCell{expanded_table.matched_time_auroc.lyapunov_divergence.humanoid.correlation_break} & \ResultCell{expanded_table.matched_time_auroc.lyapunov_divergence.breakout.correlation_break} & \ResultCell{expanded_table.matched_time_auroc.lyapunov_divergence.breakout.dynamics_scaling} \\*
PEDM & \ResultCell{expanded_table.matched_time_auroc.pedm.humanoid.correlation_break} & \ResultCell{expanded_table.matched_time_auroc.pedm.breakout.correlation_break} & \ResultCell{expanded_table.matched_time_auroc.pedm.breakout.dynamics_scaling} \\*
STI & \ResultCell{expanded_table.matched_time_auroc.sti.humanoid.correlation_break} & \ResultCellBest{expanded_table.matched_time_auroc.sti.breakout.correlation_break} & \ResultCell{expanded_table.matched_time_auroc.sti.breakout.dynamics_scaling} \\*
ACTI & \ResultCellBest{expanded_table.matched_time_auroc.acti.humanoid.correlation_break} & \ResultCell{expanded_table.matched_time_auroc.acti.breakout.correlation_break} & \ResultCell{expanded_table.matched_time_auroc.acti.breakout.dynamics_scaling} \\*
Occupancy-only & \ResultCell{expanded_table.matched_time_auroc.occupancy_only.humanoid.correlation_break} & \ResultCell{expanded_table.matched_time_auroc.occupancy_only.breakout.correlation_break} & \ResultCell{expanded_table.matched_time_auroc.occupancy_only.breakout.dynamics_scaling} \\*
ACTI+O & \ResultCell{expanded_table.matched_time_auroc.acti_occupancy.humanoid.correlation_break} & \ResultCell{expanded_table.matched_time_auroc.acti_occupancy.breakout.correlation_break} & \ResultCell{expanded_table.matched_time_auroc.acti_occupancy.breakout.dynamics_scaling} \\
\pagebreak
\toprule
Detector & \shortstack{Breakout\\Gradual drift} & \shortstack{Breakout\\Observation delay} & \shortstack{Breakout\\Action-conditioned\\dynamics} \\*
\midrule
Mahalanobis distance & \ResultCell{expanded_table.matched_time_auroc.mahalanobis_distance.breakout.gradual_drift} & \ResultCell{expanded_table.matched_time_auroc.mahalanobis_distance.breakout.observation_delay} & \ResultCell{expanded_table.matched_time_auroc.mahalanobis_distance.breakout.action_conditioned_dynamics} \\*
KDE density & \ResultCell{expanded_table.matched_time_auroc.density_kde.breakout.gradual_drift} & \ResultCell{expanded_table.matched_time_auroc.density_kde.breakout.observation_delay} & \ResultCell{expanded_table.matched_time_auroc.density_kde.breakout.action_conditioned_dynamics} \\*
Ensemble disagreement & \ResultCell{expanded_table.matched_time_auroc.ensemble_disagreement.breakout.gradual_drift} & \ResultCell{expanded_table.matched_time_auroc.ensemble_disagreement.breakout.observation_delay} & \ResultCell{expanded_table.matched_time_auroc.ensemble_disagreement.breakout.action_conditioned_dynamics} \\*
Forward reconstruction error & \ResultCell{expanded_table.matched_time_auroc.reconstruction_error.breakout.gradual_drift} & \ResultCell{expanded_table.matched_time_auroc.reconstruction_error.breakout.observation_delay} & \ResultCell{expanded_table.matched_time_auroc.reconstruction_error.breakout.action_conditioned_dynamics} \\*
State autoencoder & \ResultCell{expanded_table.matched_time_auroc.autoencoder.breakout.gradual_drift} & \ResultCell{expanded_table.matched_time_auroc.autoencoder.breakout.observation_delay} & \ResultCell{expanded_table.matched_time_auroc.autoencoder.breakout.action_conditioned_dynamics} \\*
CUSUM & \ResultCell{expanded_table.matched_time_auroc.cusum.breakout.gradual_drift} & \ResultCell{expanded_table.matched_time_auroc.cusum.breakout.observation_delay} & \ResultCell{expanded_table.matched_time_auroc.cusum.breakout.action_conditioned_dynamics} \\*
Time-series features & \ResultCellBest{expanded_table.matched_time_auroc.time_series_features.breakout.gradual_drift} & \ResultCell{expanded_table.matched_time_auroc.time_series_features.breakout.observation_delay} & \ResultCell{expanded_table.matched_time_auroc.time_series_features.breakout.action_conditioned_dynamics} \\*
DEXTER & \ResultCell{expanded_table.matched_time_auroc.dexter.breakout.gradual_drift} & \ResultCell{expanded_table.matched_time_auroc.dexter.breakout.observation_delay} & \ResultCell{expanded_table.matched_time_auroc.dexter.breakout.action_conditioned_dynamics} \\*
DEXTER+C & \ResultCell{expanded_table.matched_time_auroc.dexter_c.breakout.gradual_drift} & \ResultCellBest{expanded_table.matched_time_auroc.dexter_c.breakout.observation_delay} & \ResultCell{expanded_table.matched_time_auroc.dexter_c.breakout.action_conditioned_dynamics} \\*
Lyapunov divergence & \ResultCell{expanded_table.matched_time_auroc.lyapunov_divergence.breakout.gradual_drift} & \ResultCell{expanded_table.matched_time_auroc.lyapunov_divergence.breakout.observation_delay} & \ResultCell{expanded_table.matched_time_auroc.lyapunov_divergence.breakout.action_conditioned_dynamics} \\*
PEDM & \ResultCell{expanded_table.matched_time_auroc.pedm.breakout.gradual_drift} & \ResultCell{expanded_table.matched_time_auroc.pedm.breakout.observation_delay} & \ResultCell{expanded_table.matched_time_auroc.pedm.breakout.action_conditioned_dynamics} \\*
STI & \ResultCell{expanded_table.matched_time_auroc.sti.breakout.gradual_drift} & \ResultCell{expanded_table.matched_time_auroc.sti.breakout.observation_delay} & \ResultCellBest{expanded_table.matched_time_auroc.sti.breakout.action_conditioned_dynamics} \\*
ACTI & \ResultCell{expanded_table.matched_time_auroc.acti.breakout.gradual_drift} & \ResultCell{expanded_table.matched_time_auroc.acti.breakout.observation_delay} & \ResultCell{expanded_table.matched_time_auroc.acti.breakout.action_conditioned_dynamics} \\*
Occupancy-only & \ResultCell{expanded_table.matched_time_auroc.occupancy_only.breakout.gradual_drift} & \ResultCell{expanded_table.matched_time_auroc.occupancy_only.breakout.observation_delay} & \ResultCell{expanded_table.matched_time_auroc.occupancy_only.breakout.action_conditioned_dynamics} \\*
ACTI+O & \ResultCell{expanded_table.matched_time_auroc.acti_occupancy.breakout.gradual_drift} & \ResultCell{expanded_table.matched_time_auroc.acti_occupancy.breakout.observation_delay} & \ResultCell{expanded_table.matched_time_auroc.acti_occupancy.breakout.action_conditioned_dynamics} \\
\end{longtable}
\endgroup
\end{document}